\PassOptionsToPackage{prologue,table,xcdraw}{xcolor}
\documentclass[sigconf, nonacm,pdfa]{acmart}

\usepackage[a-2b]{pdfx} % PDF/A compliance See: https://github.com/fanju1984/pvldb-pdfa-resources

\usepackage{enumitem}
\usepackage{balance}
\usepackage{float}
\usepackage{pgfplots}
\pgfplotsset{compat=1.18} 
\usepackage{booktabs}
\usepackage{algorithm}
\usepackage[noend]{algpseudocode}
\usepackage{afterpage}
\usepackage[english]{babel}
\usepackage[most]{tcolorbox}
\usepackage[framemethod=tikz]{mdframed} % Colored box
\usepackage{fancyvrb}

\newcommand{\Figref}[1]{Fig.~\ref{#1}}
\newcommand{\Tabref}[1]{Tab.~\ref{#1}}
\newcommand{\Secref}[1]{Sec.~\ref{#1}}
\newcommand{\Algref}[1]{Algo.~\ref{#1}}

\newcommand{\myparagraph}[1]{\vspace{.1cm} \noindent \textbf{#1}}
\newcommand{\myparagraphemph}[1]{\noindent \emph{#1}}
\newcommand{\hide}[1]{}

\newtheorem{mdexample}{\hspace{5pt}Example}

\newtheorem*{mdexample*}{Example}
\newenvironment{example*}{\begin{mdframed}[backgroundcolor=teal!12, roundcorner=10pt, linewidth=0pt, innertopmargin=1pt, innerbottommargin=5pt, skipabove=9pt, skipbelow=3pt]\begin{mdexample*}}{\end{mdexample*}\end{mdframed}}

\newcommand\vldbdoi{10.14778/3665844.3665857}
\newcommand\vldbpages{2279 - 2292}% issue-specific
\newcommand\vldbvolume{17}
\newcommand\vldbissue{9}
\newcommand\vldbyear{2024}
\newcommand\vldbauthors{\authors}
\newcommand\vldbtitle{\shorttitle} 
\newcommand\vldbavailabilityurl{https://github.com/penfever/ArcheType}
\newcommand\vldbpagestyle{empty} 

\begin{document}
\title{ArcheType: A Novel Framework for Open-Source \\ Column Type Annotation using Large Language Models} %(Flavors:  Systems)}

\author{Benjamin Feuer}
\orcid{0000-0002-7938-542X}
\affiliation{%
  \institution{New York University}
  %\country{United States of America}
}
\email{bf996@nyu.edu}
\author{Yurong Liu}
\affiliation{%
  \institution{New York University}
  %\country{United States of America}
}
\email{yurong.liu@nyu.edu}
\author{Chinmay Hegde}
\affiliation{%
  \institution{New York University}
  %\country{United States of America}
}
\email{chinmay.h@nyu.edu}
\author{Juliana Freire}
\affiliation{%
  \institution{New York University}
  %\country{United States of America}
}
\email{juliana.freire@nyu.edu}

\begin{abstract}
Existing deep-learning approaches to semantic column type annotation (CTA) 
have important shortcomings: they rely on semantic types which are fixed at 
training time; require a large number of training samples per type; 
incur high run-time inference costs; and their performance can degrade when evaluated 
on novel datasets, even when types remain constant. Large language models have exhibited 
strong zero-shot classification performance on a wide range of tasks and in this paper we explore their use for CTA. 
We introduce ArcheType, a simple, practical method for context sampling, prompt serialization, model querying, and label remapping, which enables large language models to solve CTA problems in a fully zero-shot manner. 
We ablate each component of our method separately, and establish that improvements to context sampling and label remapping provide the most consistent gains. 
ArcheType establishes a new state-of-the-art performance on zero-shot CTA benchmarks (including three new domain-specific benchmarks which we release along with this paper), and when used in conjunction with classical CTA techniques, it outperforms a SOTA DoDuo model on the fine-tuned SOTAB benchmark. 
ArcheType establishes a new state-of-the-art performance on zero-shot CTA benchmarks (including three new domain-specific benchmarks which we release along with this paper), and when used in conjunction with classical CTA techniques, it outperforms a SOTA DoDuo model on the fine-tuned SOTAB benchmark. 
\end{abstract}

\maketitle

\pagestyle{\vldbpagestyle}
\begingroup\small\noindent\raggedright\textbf{PVLDB Reference Format:}\\
\vldbauthors. \vldbtitle. PVLDB, \vldbvolume(\vldbissue): \vldbpages, \vldbyear.\\
\href{https://doi.org/\vldbdoi}{doi:\vldbdoi}
\endgroup
\begingroup
\renewcommand\thefootnote{}\footnote{\noindent
This work is licensed under the Creative Commons BY-NC-ND 4.0 International License. Visit \url{https://creativecommons.org/licenses/by-nc-nd/4.0/} to view a copy of this license. For any use beyond those covered by this license, obtain permission by emailing \href{mailto:info@vldb.org}{info@vldb.org}. Copyright is held by the owner/author(s). Publication rights licensed to the VLDB Endowment. \\
\raggedright Proceedings of the VLDB Endowment, Vol. \vldbvolume, No. \vldbissue\ %
ISSN 2150-8097. \\
\href{https://doi.org/\vldbdoi}{doi:\vldbdoi} \\
}\addtocounter{footnote}{-1}\endgroup

\ifdefempty{\vldbavailabilityurl}{}{
% \vspace{.3cm}
\begingroup\small\noindent\raggedright\textbf{PVLDB Artifact Availability:}\\
The source code, data, and/or other artifacts have been made available at \url{\vldbavailabilityurl}.
\endgroup
}

\section{Introduction}
\label{sec:intro}

The goal of semantic column type annotation (CTA) is to associate each column of a relational table with one among several pre-defined semantic types that go beyond atomic types such as string, integer, or Boolean. CTA is a useful computational primitive in numerous settings, including data cleaning, where detection, correction, and transformation are performed using rules based on data types~\cite{raman2001potter, kandel2011wrangler},
and schema matching for data discovery, where the semantic type can be used to constrain the search for matching attributes~\cite{khatiwada2023santos,chu2019data}. Beyond being useful from a computational standpoint, efficient methods for CTA can also enable democratization of access to large, well-curated datasets by reducing labeling costs.

\myparagraph{Learning-Based CTA.} Recent approaches to CTA have increasingly been based on learning-based techniques. Deep learning approaches rely on the availability of large training corpora of columns annotated with their semantic types to train a deep neural network from scratch that can perform CTA on new, unseen columns of relational tables~\cite{sherlockpaper,zhang2020sato}. Fine-tuned models, on the other hand, rely on pre-trained transformer-based language models (LMs) such as BERT~\cite{transformer} and fine-tune them for the specific task of CTA~\cite{TURLpaper,doduopaper}. Learning-based  approaches have been shown to be effective for identifying generic types for which there exists sufficient training data. For example, Sherlock~\cite{sherlockpaper} was trained on over 675,000 columns retrieved from the VizNet corpus to recognize 78 semantic types from DBpedia~\cite{auer2007dbpedia} such as album, city, plays, or birth place. 
However, these approaches exhibit important limitations with respect to distribution shift, the need for large volumes of training data, and the cost involved in supporting rare type.
First and foremost, their performance degrades substantially when evaluated against test datasets that have been acquired from different sources \emph{even when their column types match closely}. This problem is sometimes called \emph{distribution shift}~\cite{quinonero2008dataset}.
An important desideratum of deep learning models is that they exhibit predictable model behavior under natural distribution shifts, i.e., when evaluation data which differs from the data on which a model was trained due to natural factors.
% , in contrast with human vision
However, recent works show that the vast majority of standard deep models for image classification perform significantly worse under natural shifts~\cite{Hendrycks2019BenchmarkingNN, Miller2021AccuracyOT, Recht2019DoIC}.

We posit that the same phenomenon occurs in closed-set deep learning models for CTA.
%column type annotation. 
%
%\vspace{-.3cm}
\begin{example*}
Suppose we fix a given column type $\texttt{location}$ and that our pre-training distribution is sourced from NYC Open Data~\cite{nyc_opendata}. Then we might see entries like \texttt{Broadway}, \texttt{SoHo}, \texttt{Jamaica},
which are locations in New York City. But if we use this model 
to perform CTA on a dataset from the Brazilian Dados Abertos~\cite{dadosabertos}, it is unlikely to assign the $\texttt{location}$ label to
$\texttt{Corcovado}$ and $\texttt{Lapa}$, which are locations in Rio de Janeiro.
\end{example*}
As a simple empirical validation of this problem, we compared the performance of the fine-tuned DoDuo CTA model~\cite{doduopaper},
on the Schema.Org Table Annotation Benchmark (SOTAB)~\cite{sotab}. We use the DoDuo variant pretrained on the similar VizNet dataset~\cite{viznet}, reusing CTA labels from that benchmark wherever possible. We find that performance declines over 60\% (from 84.8\% to 23.8\%).

\begin{figure*}[t]
    \centering
    \includegraphics[width=.75\textwidth]{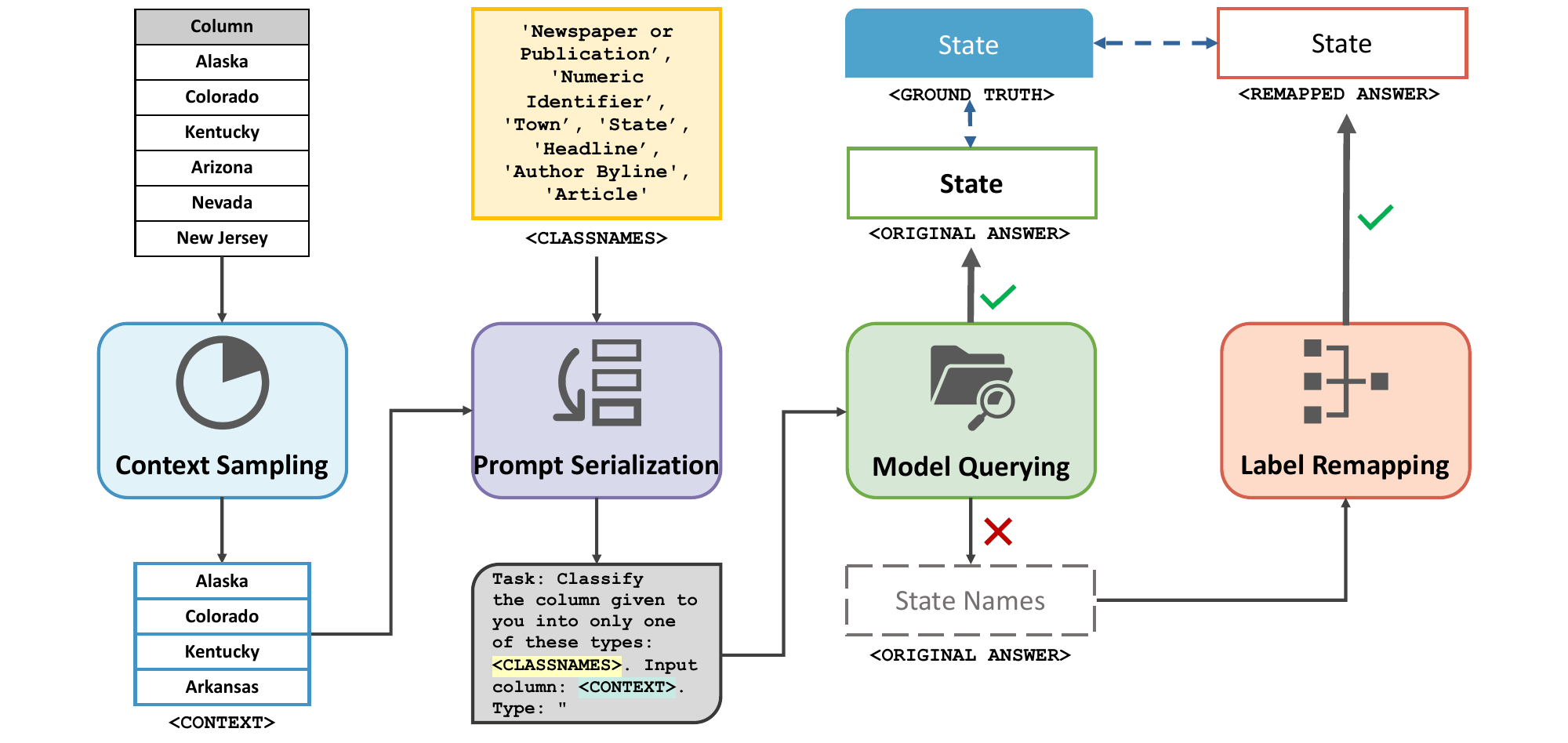}
    \vspace{-.3cm}
    \caption{\sl\textbf{ArcheType: a four-stage method for column type annotation. } \sl  (1) In the Context Sampling stage, an algorithm selects a few representative samples from a column. (2) In the Prompt Serialization stage, the context and instruction string are serialized in a model-specific, token-efficient manner. (3) The prompt is input to a LLM in the Model Querying stage. (4) If the output of the LLM is not one of the allowable categories, the Label Remapping stage assigns the model output to a class. }
    \Description[ArcheType: a four-stage method for column type annotation.]{In the Context Sampling stage, an algorithm selects a few representative samples from a column. (2) In the Prompt Serialization stage, the context and instruction string are serialized in a model-specific, token-efficient manner. (3) The prompt is input to a LLM in the Model Querying stage. (4) If the output of the LLM is not one of the allowable categories, the Label Remapping stage assigns the model output to a class.}
  \label{fig:llmcta}
  \vspace{-.3cm}
\end{figure*}

Even if existing models did not struggle under shift, their utility is still constrained by the fact that \emph{label sets have to be specified at training time}. However, real-world data is vast, and pre-trained type labels rarely map cleanly to categories of interest in newly-encountered datasets; in many scenarios datasets do not have a schema which fit neatly into these pre-trained types. 

\begin{example*}
Consider the NYC Open Data repository~\cite{nyc_opendata} which contains thousands of datasets published by NYC agencies and includes NYC-specific semantic types such as public schools, agencies, parks, and boroughs. 
As point of reference regarding the specificity of this collection, \citet{D4paper} computed the overlap between the contents of datasets in
NYC Open Data and word vectors trained with GloVe (which uses Wikipedia as a source) and found that GloVe covers only 8\% of the terms in the collection.
We note that existing ontologies and taxonomies such as DBpedia~\cite{auer2007dbpedia} define generic types that encompass the NYC-specific types, gor example, a high school can be classified as $\texttt{EducationalInstitution}$. However, this %semantic 
type includes many institution types that are not public schools, such as colleges, medical centers and libraries. If we use this semantic type to find tables to augment information about NYC high schools, many irrelevant tables would be retrieved.
\end{example*}

Further, \emph{training a model to recognize new types is both time-consuming and costly} as it requires the acquisition of labeled data and the training of new deep models. This can severely limit the applicability of learning-based approaches~\cite{sherlockpaper,doduopaper,TURLpaper} to long-tail and rare types, which can negatively affect downstream applications.

Moreover, the \emph{volume of training data required by modern CTA models is substantial}. For example, Sherlock~\cite{sherlockpaper} was trained on over 675,000 columns retrieved from the VizNet corpus to recognize 78 semantic types from DBpedia~\cite{auer2007dbpedia}.
Over 397,000 tables were used for training versions of the current state-of-the-art DoDuo~\cite{doduopaper}. This imposes high data cleaning and labeling costs which can be oppressive, particularly for infrequent classes.

\myparagraph{Using LLMs for CTA.}
As a silver lining, the recent dramatic advances in generative large language models (LLMs) open the opportunity to address these challenges and create robust models for
a broad set of semantic types without requiring large volumes of labeled data.
LLMs are trained over a very large and diverse corpus and they are thus able to \emph{accumulate knowledge that covers a plethora of semantic types}. Furthermore, these modes have the capability to perform \emph{in-context learning}, where the label set can be specified as user-defined context during inference time, making it possible to \emph{perform open-set classification even for rare types}.
\begin{example*}
For example,  when presented with  $\texttt{Stuyvesant}$, 
    GPT-3.5-Turbo learns in-context that it is being asked to do classification and asserts it is a $\texttt{High School in New York City}$.
\end{example*}
This capability enables \emph{zero-shot} CTA as well as the generation of labels that can be used to fine-tune models for domain-specific types.  LLMs have also been shown to perform much better than other learning-based models under distribution shift~\cite{Radford2021LearningTV}, opening the possibility for the creation of robust CTA models.

\myparagraph{Our contributions.} In this paper, we take several steps towards establishing the effectiveness and limitations of LLMs for CTA.
We discuss the challenges involved in using LLMs for CTA and systematically delineate the different components required to perform CTA using LLMs: sampling the data context, prompt serialization, model querying, and label remapping (illustrated in \Figref{fig:llmcta}). We propose novel methods for these components and assess their effectiveness. 

We also explore the impact of these components on two different modes of operation: (a) using existing LLMs for zero-shot CTA and (b) fine-tuning LLMs for CTA based on a training set of labeled column types.
For both modes of operation, we report a series of results for \emph{open-source LLMs}. As a basis of comparison we also study and report the performance of a closed-source LLM (the GPT family).
However, we emphasize open-source LLMs in our work, since closed-source models are not transparent: since we do not know how they were constructed, it may be difficult to understand their behavior;  and since closed-source LLMs are constantly updated, reported results cannot be reproduced~\cite{chen2023analyzing}.

We perform a detailed evaluation of our approach against state-of-the-art learning-based CTA systems~\cite{doduopaper,TURLpaper,sherlockpaper} as well as a new zero-shot approach~\cite{kayali2023chorus}. We use established benchmarks~\cite{viznet,efthymiou2017matching,chen_learning_2019} and the SOTAB benchmark, which was designed for comparing the performance of annotation systems on CTA tasks~\cite{sotab}. However, we observe that these benchmarks are primarily composed of well-known semantic types drawn from widely-used ontologies and taxonomies. To explore the breadth of LLM subject knowledge as well as how LLM-based CTA performs for a wide range of types (including rare, domain-specific types with novel characteristics), we also introduce three new benchmark datasets for CTA, described in \Secref{sec:new-benchmarks}.

Our main contributions can be summarized as follows:
\begin{enumerate}[leftmargin=*,nosep]
\item We introduce \textbf{ArcheType}, an open-source CTA framework centered around large language models, which leverages their strengths, adapts to their limitations, and is compatible with both open-source and closed-source LLMs.
\item We enumerate four essential components for any LLM-based CTA (LLM-CTA) approach: sampling, serialization, querying, and label remapping. We propose new approaches for context sampling and label remapping, and demonstrate their importance to the overall accuracy of LLM-CTA  (\Secref{sec:archetype-method}).
\item We introduce three new zero-shot CTA benchmarks that cover a range of domain-specific schemas and attribute types (\Secref{sec:new-benchmarks}).
\item Through a detailed experimental evaluation (\Secref{sec:experiments}), we show that ArcheType achieves strong fine-tuned performance and state-of-the-art zero-shot performance on a large and diverse suite of benchmarks, while requiring far less tabular data for both training and inference than existing methods (\Secref{sec:main-results}).
%\vspace{-.15cm}
\end{enumerate}

\vspace{-.2cm}
\section{Background: Foundation Models}

The term \textbf{foundation model} applies to large machine learning models that are pre-trained on vast amounts of raw data to capture a wide range of knowledge, and then fine-tuned on more specific tasks or datasets~\cite{bommasani2021opportunities}. In the case of large language models (LLMs), the pre-training objective is autoregressive; the model is tasked with predicting the next word in a sequence based on the context provided by the preceding words. 
The scale of LLMs results in new emergent capabilities, and their effectiveness across a multitude of tasks incentivizes the use of foundation models as a starting point (or 
%drop-in 
replacement) for fine-tuning task-specific models. However, this last step must be done with care since the defects of the foundation model are inherited by all the adapted models downstream~\cite{bommasani2021opportunities}.

\vspace{-.3cm}
\subsection{LLMs and Tabular Data}
\label{sec:llm-classif}

The development of LLMs has largely been driven in the context of NLP tasks as question-answering, logical inference, and word disambiguation. 
Recent efforts based on instruction-following, such as~\cite{Ouyang2022TrainingLM} and~\cite{Chung2022ScalingIL}, have demonstrated that fine-tuning foundational LLMs on a carefully curated corpus of prompt-response pairs is an effective strategy for more generic classification tasks. However, these approaches focus on natural language datasets that have small label sets, clean labels, and balanced classes. 

There have been only a handful of attempts to apply LLMs to tasks that are germane to tabular data. Recently, \citet{Hegselmann2022TabLLMFC} proposed a LLM-based framework for few-shot classification of tabular data and experimented with different strategies to design the prompt. They showed that their approach can outperform state-of-the-art (SOTA) neural models both in the zero- and few-shot settings.
\citet{Narayan2022CanFM} outline a vision for leveraging LLMs for data management tasks and show that LLMs using few-shot and zero-shot approaches can achieve SOTA performance for entity matching, data imputation, and error detection.

\vspace{-.15cm}
\subsection{LLMs for Zero-Shot CTA}
\label{sec:llms-for-zero-shot-cta}

\begin{table}[t]
\caption{\sl\textbf{Cost of CTA benchmarking with GPT. } \sl Approximate cost to perform CTA over the 15,040 column test set of the SOTAB dataset varying the table serialization \emph{Method}
(column for column-at-once or table for table-at-once); the number of context samples \emph{\#Smp.} drawn per column;  the percentage \emph{\%\>k} of serialized prompts whose tokenized length is estimated to exceed a context window of size k.} 
\vspace{-.3cm}
\resizebox{0.95\linewidth}{!}{
\begin{small}
\begin{tabular}{lrrrrr}
\textbf{Method} & \multicolumn{1}{l}{\textbf{\# Smp.}} & \multicolumn{1}{l}{\textbf{\% \textgreater 1k}} & \multicolumn{1}{l}{\textbf{\% \textgreater 4k}} & \multicolumn{1}{l}{\textbf{\% \textgreater 16k}} & \multicolumn{1}{l}{\textbf{App. USD Cost}} \\
\midrule
column          & 3                                        & 01.0\%                                               & 00.1\%                                             & 00.0\%                                                & \$7.85                                                 \\
column          & 10                                       & 06.5\%                                             & 00.3\%                                             & 00.0\%                                                & \$12.54                                                \\
column          & 20                                      & 13.0\%                                             & 01.3\%                                             & 00.1\%                                              & \$19.97                                                \\
column          & 100                                     & 93.0\%                                              & 15.1.\%                                            & 01.0\%                                                & \$90.38                                                \\
column          & 1000                                     & 100.0\%                                             & 100.0\%                                             & 56.5\%                                             & \$1,072.06                                                \\
table           & 10                                      & 89.8\%                                            & 58.8\%                                            & 25.8\%                                             & \$763.28  
\end{tabular}
\end{small}
}
\vspace{-.3cm}
\label{tab:sampling-cost}
\end{table}

As discussed in \Secref{sec:intro}, LLMs present new opportunities to derive robust models for CTA that can handle a broad set of classes at a much lower cost than existing learning-based methods.
Two recent approaches have been proposed that leverage OpenAI's GPT for zero-shot and few-shot CTA~\cite{korini2023column, kayali2023chorus}. 
These methods do not require model training, and apply open-vocabulary labels either from parametric memory~\cite{kayali2023chorus}, or from options provided at test time~\cite{korini2023column}.

The promise of such a direction is clear, but existing implementations have important limitations. 
Both~\cite{korini2023column, kayali2023chorus}, which are to the best of our knowledge  the only existing works on zero-shot CTA, rely on closed-source models (see discussion below).
They also require access to the entire table at test time to achieve their best performance, which in practice can be expensive for private models.
\begin{example*}
\Tabref{tab:sampling-cost} demonstrates that the cost to evaluate the SOTAB test set (assuming sampling with replacement) scales poorly for table-at-once methods, and over 25\% of the prompts exceed the maximum possible context window. The cost is also high for column-at-once methods when a large sample is used.
\end{example*}
Since these methods are highly sensitive to sample size, it is important to devise strategies that are sample-efficient. However, only simple random sampling and first-k-rows sampling methods have been
explored for LLM-based CTA. 
Note that while these methods are costly on closed-source models, they can be impractical on open-source models, owing to their limited context windows.

A distinct line of inquiry studied by \citet{tu_unicorn_2023} treats CTA as one example of a family of matching tasks in data integration, and is able to perform zero-shot binary matching on CTA instances.

\vspace{-.15cm}
\subsection{Open vs. Closed-Source LLMs}
We consider a model \textbf{open-source} if, and only if, sufficient specifics of model design have been published to reproduce the architecture, checkpoints with pre-trained weights have been released  and the contents of the pre-training corpus are available for inspection. 
The advantages of utilizing open-source models are explainability, reproducibility, and reduced cost, while the drawbacks are performance and limited context length.

\myparagraphemph{Explainability.} The architectures of most closed-source models are not known to the public; nor is it known how much prompt engineering and behind-the-scenes modification of the model output is being conducted. The specifics of the data on which these models are trained is also unknown. These facts make it difficult to provide rigorous explanations for the behavior of closed-source models.

\myparagraphemph{Reproducibility.} As noted recently, results from closed-source models are non-reproducible, non-deterministic, and cannot be ablated with respect to the model architecture or dataset, all of which makes them unreliable for reproducible research~\cite{pradeep2023rankvicuna, rogers-etal-2023-closed}.

\myparagraphemph{Cost.} As closed-source models charge by the token, the cost incurred by any solution which relies on them can be considerable. Open-source models, by contrast, require computational resources to host and expertise to maintain.

\myparagraphemph{Performance.} As of this writing, the best open-source models underperform the best closed-source models across a wide range of benchmarks~\cite{Liang2022HolisticEO}. The causes of this performance gap are not fully understood, as large language models tend to exhibit unpredictable phase transitions as a function of scale. These transitions can lead to sudden leaps in performance on standard benchmarks~\cite{Power2022GrokkingGB}.

\myparagraphemph{Context length.} The open-source large language models in common use at the time of this paper have context windows ranging from 512 to 2048 tokens~\cite{Chung2022ScalingIL, Touvron2023LLaMAOA} (typically between 375 and 1500 words, if the string is English). If the string is in a different language or is largely numeric, however, the tokenization process tends to be approximately 2-4x times less efficient, since standard tokenization schemes employed by such models tend to handle unicode inefficiently~\cite{Sennrich2015NeuralMT}. Both phenomena are common in real-world tabular data. Closed-source models are less constrained (GPT-3.5 allows over 16,000 tokens at the time of this writing).

\vspace{-.1cm}
\section{ArcheType: Methods and System}
\label{sec:archetype-method}

\myparagraph{Formal Model of LLM-CTA.} Consider a table $T$ with $t$ columns and $r$ rows. We denote each column $C \in T$ as a function which maps row indices to strings; i.e., for $0 <= i < t$, we have $C_i : \mathbb{N} \rightarrow \Sigma_*$, where $i$ is the column index. Here, $\Sigma_*$ is the set of all possible strings, $\Sigma_{C_i}$ is the set of all strings found in column $C_i$, $\Sigma_{C_i} \subset \Sigma_*$ $\forall i$, with any individual string $\sigma \in \Sigma_{C_i}$. We make no further assumptions; $C_i$ may include a column name, and $T$ may contain an additional metadata field. However, neither of these properties is required to exist, and so we do not include them in our analysis. Many of our methods rely on a sample of \textit{unique} values sampled from the column, $U_i := \texttt{unique}(|\Sigma_{C_i}|)$.
We explore two LLM-based approaches for CTA: fine tuned and zero shot.

\begin{definition}[Fine-tuned LLM-CTA]
Let $L \subseteq \Sigma_*$ denote a label set; these are our column types to be annotated. Standard CTA assumes a fixed cardinality for this label set, indexed by a variable we call $j$.\footnote{In existing benchmarks, $j$ can be anywhere from 10 to 300~\cite{korini2023column}} 
Given the above definitions, we define fine-tuned single-label $CTA \subset T \times L$ as a relation between tables and labels:
\vspace{-.15cm}

\begin{equation}
    \forall C, \exists l_j \mid (C_i,l_j) \in CTA
\end{equation}

\noindent We seek a generative method $M : \Sigma_* \rightarrow \Sigma_*$ that comes closest to satisfying the following properties:
\begin{equation}
    M(\sigma, L) \in L, \forall C \in T, M(\sigma, L) \in CTA
\end{equation}

\noindent i.e., the model requires a single string as input and generates a label in $L$ that correctly represents the type of $C$.

\end{definition}

\begin{definition}[Zero-shot LLM-CTA]

The definition of zero-shot LLM-CTA is identical to that of fine-tuned, except that: in a zero-shot setting, the number of rows $r$ is presumed to be small enough to preclude the possibility of fine-tuning a model; $L$ is chosen at test-time; and  it is possible to define multiple values of $L$ for one $T$.

 \end{definition}

\vspace{-.25cm}
\subsection{Elements of LLM-CTA Methods}
\label{sec:elements-of-llm-cta}
  
We observe that any LLM-CTA method must provide solutions to four problems: context sampling, prompt serialization, model querying, and label remapping. Individually, each is necessary for LLM-CTA; collectively, they are sufficient.
By considering and ablating approaches to each of these problems separately, we designed ArcheType,  a LLM-CTA framework  which generalizes to a wide range of architectures, including popular open-source models. 
\Figref{fig:llmcta} provides an overview of ArcheType and in the remainder of this section, we describe its components in detail.

\myparagraph{Context Sampling.} As of this writing, all SOTA large language models (LLMs) are transformer-based~\cite{transformer}. By design, transformers have a hard scaling limit over which their dense attention can be applied, sometimes called a \textit{context window}, $W$. Given a context $C$ and a set of labels $L$, if $|C| + |L| > W$, a representative sample must be selected. From a practical standpoint, the context window sizes of contemporary LLMs are small enough that this event takes place quite frequently, e.g., \cite{kayali2023chorus} and \cite{korini2023column} use simple random sampling and first-k-rows sampling, respectively. We introduce a new sampling method in \Secref{sec:context-sampling} and provide ablation studies in \Secref{sec:ablations-context-sampling}.

\myparagraph{Prompt Serialization.} SOTA LLMs require prompts, or priors, to complete. Prompt serialization (or prompt engineering) is the process of transforming raw context into a prompt. Of the four components we consider here, this one has received the most attention in the existing literature; the methods introduced by \cite{korini2023column, kayali2023chorus} are largely focused on improvements to prompt serialization. In \Secref{sec:ablations-prompt-ser}, we ablate prompt serialization, independent of other components, and conclude prompt engineering should be treated as a hyperparameter rather than as a methodological contribution -- we describe this approach in \Secref{sec:prompt-ser}. When considering a range of model architectures, we find that any reasonable serialization method is about as likely to produce a good result as any other.

\myparagraph{Model Querying.} Model selection and querying is another important element of LLM-CTA. The method must correctly submit a query to some large language model(s) chosen in advance, and it must retrieve and process the response. This query may be processed on a local machine or via an API. This, too, has not been ablated in prior work. While future work may attempt to train a generative large language model from scratch specifically for this task, \cite{korini2023column, kayali2023chorus} use GPT, and only GPT. As part of our study, we present ablations on architectures across a range of open-source models as well as GPT (\Secref{sec:ablations-model-querying}) and find that no model dominates.

\myparagraph{Label Remapping.} All LLMs sometimes produce responses which do not match with any of the labels provided in the prompt, i.e., $\sigma_L \notin L$. Label remapping is a form of error correction which remaps an unbounded LLM output space to a limited set of labels. \citet{kayali2023chorus} use an embedding-based method called anchoring to remap labels, whereas \citet{korini2023column} use a dictionary lookup. As the latter approach is not compatible with zero-shot LLM-CTA, we ablate only the former approach, along with two other baselines, and develop CONTAINS+RESAMPLE (\Secref{sec:label-remap}), an algorithm which outperforms the baselines across model architectures. We ablate our choice of remapping method in \Secref{sec:ablations-label-remapping}.

\vspace{-.15cm}
\subsection{Context Sampling}
\label{sec:context-sampling}

CTA approaches using deep learning face severe data requirement challenges in settings that require (very) large tables and open label sets. To address these challenges, we introduce a new approach which we call \textbf{context sampling} and outline in
\Algref{alg:context_sampling}. Given the unique values of a target column $U_i$ and a target sample size $\phi$, we seek to construct the representative sample $S$ that best summarizes the column. While it is possible in LLM-CTA to have $\phi$ vary by column, in this paper we consider the setting where $\phi$ is fixed in advance and consistent across all columns.

In the simplest case, we have $|U_i| \geq \phi$, and $S$ is drawn without replacement from a distribution whose construction is described below. If $|U_i| < \phi$, then $S$ is drawn with replacement instead.

In the fine-tuned setting, we find it is beneficial to add %more
features to the context window, affecting both sampling and serialization. The features we utilize are described later in this section, and are sampled as described in \Algref{alg:context_sampling}.
The context sample is then serialized and embedded into a prompt which is passed to the LLM, the format of which follows from recent works such as~\cite{Muennighoff2022CrosslingualGT} and~\cite{Chung2022ScalingIL}.

\myparagraph{Context Sampling in ArcheType.} The probability distribution over $U_i$ from which we sample is weighted according to an importance function $f$. The probability of selecting an element \( \sigma \) from \( U_i \) under \( P_{U_i} \) is given by:
\vspace{-.3cm}
\[ P(\sigma) = \frac{f(\sigma)}{\sum_{j \in U_i} f(\sigma_j)}. \] 

\noindent We utilize two importance functions in ArcheType. For the American Stories (amstr) benchmark described in \Secref{sec:new-benchmarks}, we find that an importance function which prioritizes unique samples that include any target class name is most effective; $f(\sigma) = 1$ if, for any $l_j \in L$, $l_j \subset \sigma$, else $f(\sigma) = 0.1$. Note that this function does not require the ground truth label of any particular sample, only the entire label set, which is a required input for CTA. 

For all other benchmarks, the importance function $f$ is string length – our experiments showed that long strings lead to better results. One possible reason is that longer strings are more likely to contain useful information than shorter ones. While an extensive ablation of the choice of importance function is beyond the scope of this paper, we note that ArcheType users (subject matter experts) can define importance functions suitable for their applications.

\begin{algorithm}[t]
\begin{algorithmic}[1]
\Procedure{Context-sample}{$T, i, S, \phi, P, SS, TN, E$} \Comment{$T$: A table, $i$, a target column index, $S$, a context sample to be returned, $\phi$: A hyperparameter (number of samples), $P$: a valid probability function, $SS$: summary statistics, $TN$: table name, $OC$: other columns, $E$: extended context flag}
\State $U_i \gets \text{UNIQUE}(T_i)$, $S \gets \emptyset$
\If{E} $S \gets \text{SS}(T_i) + \text{TN}(T)$ \EndIf
\While{$|S| \leq \phi$}
\State $S \gets S + \sigma \sim P_{U_i}$\Comment{Drawn without replacement}
\EndWhile
\If{$(|S| < \phi) \land E$}
\For{$j \in T, j \neq i$}
\State $S \gets S + T_j[0]$
\If{$|S| \geq \phi$} BREAK \EndIf
\EndFor
\Else
\While{$|S| \leq \phi$}
\State $S \gets S + \sigma \sim P_{U_i}$\Comment{Drawn with replacement}
\EndWhile
\EndIf
\State return $S$
\EndProcedure
\end{algorithmic}
%\vspace{13pt}
\caption{\sl\textbf{Context sampling. } \sl  
%The procedure accepts as input 
Given a table $T$, a valid probability function $P$, and optional additional features, produce a context sample $S$ of the appropriate size. If $|U_i| \rightarrow \infty$, methods like~\cite{distinctvalues} can be used to derive a finite-size $U_i$.} 
%The secondary features we use are described under the heading Feature Selection in \Secref{sec:context-sampling}}
\label{alg:context_sampling}
\end{algorithm}

There are challenges in the implementation of context sampling, including \emph{low variance (degenerate) data}
$U_i \ll o(1)$ and 
\emph{high variance data} 
$U_i \gg \phi$.
Each of these situations merits discussion.

\myparagraph{High variance.} In this case, helpful context may be lost in a limited sample. This phenomenon may explain why increasing the size of the context sample tends to improve model performance. However, the improvements are slight, suggesting an exponential scaling of data demands, similar to those noted by \cite{sutton2019bitter}. 
\myparagraph{Low variance.} CTA can easily become unsolvable for \emph{low-variance} or, in the extreme case, \emph{degenerate columns}.
\begin{example*}
Consider a column $C_d$ such that 
$$\forall k \in U_i, \Sigma_{U_{i_k}} =``0"$$
and a label set \texttt{L = number, integer, quantity}.
There exists no unique $\sigma_{L_j}$ such that 
$
CTA(C_d, L_j) = \sigma_{L_j}
$.
\end{example*}

In some cases, we find that incorporating additional metadata (such as the filename of the table) can help with the classification task, but in other cases, we found that it simply biases the LLM to parrot back portions of the input string.

\myparagraph{Feature Selection.}
In context sampling, \textit{feature selection} refers to what aspects of the original data we choose to include in the context. In all of our experiments, our first feature is the context sample itself (CS). We also experiment with including the \textit{file name (FN)} of the table, used by \cite{TURLpaper}, \textit{summary statistics (SS)}, used by \cite{sherlockpaper}, and samples from \textit{other columns (OC)}, used by \cite{doduopaper}. 

\myparagraph{Summary statistics (SS).} SS feature selection proceeds as follows:

\begin{itemize}[leftmargin=*]
    \item We select statistics which support fast, accurate sketching.
    \item We select measures of center and spread which can provide additional information about missing column values.
\end{itemize}

The list of summary statistics included in our fine-tuned models was:
\text{standard deviation}, \text{average}, \text{mode}, \text{median}, \text{max}, \text{min}.
When the summary statistic is a floating-point value, we round it to two decimal places. When it is an integer, we exclude the decimal place. When all sampled values are numeric, the statistics are computed with respect to the individual \textit{column values}. When any sampled value is non-numeric, the statistics are computed with respect to \textit{column value lengths}.

We postulate that these statistics are useful because they help the model disambiguate between numeric column samples by preserving information about overall trends in the column. However, we focused on simple-to-calculate statistics and did not extensively ablate our choices; in future work we plan to explore this aspect.

\myparagraphemph{Other columns.} First, we take as many unique samples as are available from the target column. Then, we fill the remaining context length with an equal number of samples from each other column. We label samples from other columns with an index number in order to identify from which column they originated. Performing this improves fine-tuned performance, but has a negative effect on zero-shot performance; see \Figref{fig:context_type_ablations}. This is likely because the LLM cannot distinguish inter-column from intra-column values without the presence of learned special characters as provided in~\cite{TURLpaper, doduopaper}.

\begin{figure}[t]
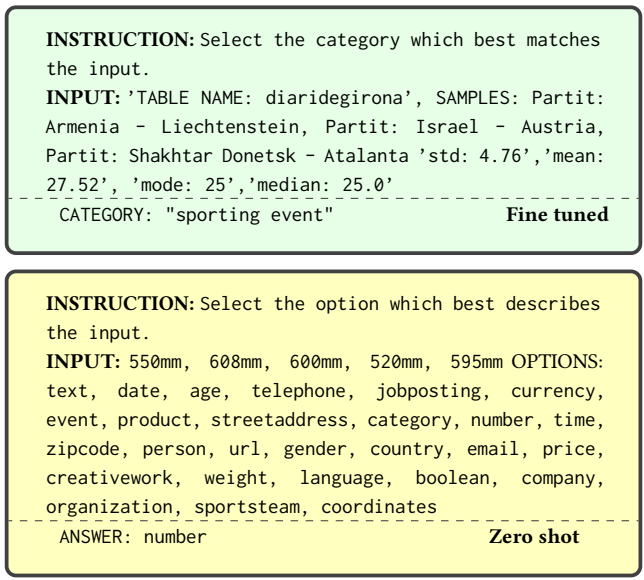

     \centering
\begin{tcolorbox}[colback=green!10!white,colframe=white!25!black]
\begin{small}
\textbf{INSTRUCTION:} \texttt{Select the category which best matches the input.} \\
\textbf{INPUT:}
\texttt{'TABLE NAME: diaridegirona',
SAMPLES: Partit: Armenia - Liechtenstein,
Partit: Israel - Austria,
Partit: Shakhtar Donetsk - Atalanta
'std: 4.76','mean: 27.52', 'mode: 25','median: 25.0'}
\end{small}
\vspace{-.25cm}
\tcblower
\begin{small}
%\begin{verbatim}
%CATEGORY: "sporting event" 
%\end{verbatim}
\vspace{-.3cm}
\begin{tabular}{lp{3.3cm}}
    \texttt{CATEGORY: "sporting event"} &  \hfill \textbf{Fine tuned} 
\end{tabular}
\end{small}
\end{tcolorbox}

%\vspace{-.3cm}

\centering
\begin{tcolorbox}[colback=yellow!25!white,colframe=white!25!black]
\begin{small}
\textbf{INSTRUCTION:} \texttt{Select the option which best 
describes the input.}\\
\textbf{INPUT:} \texttt{550mm, 608mm, 600mm, 520mm, 595mm} 
OPTIONS: \texttt{text,  date,  age,  telephone,  jobposting,  
currency,  event,  product,  streetaddress,  
category,  number,  time,  zipcode,  person,  
url,  gender,  country,  email,  price,  
creativework,  weight,  language,  boolean,  
company,  organization,  sportsteam,  
coordinates}
\end{small}
\vspace{-.25cm}
\tcblower
\begin{small}
\vspace{-.3cm}
\begin{tabular}{lp{4.6cm}}
    \texttt{ANSWER: number} &  \hfill \textbf{Zero shot} %\vspace{-.4cm}
\end{tabular} 
%\begin{verbatim}
%ANSWER: number
%\end{verbatim}
\end{small}

\end{tcolorbox}
%\vspace{-.3cm}
        \vspace{-.3cm}
     \caption{\sl\textbf{Examples of ArcheType fine-tuned (top) and zero-shot (bottom) prompting.} \sl }
     \Description[Examples of ArcheType fine-tuned (top) and zero-shot (bottom) prompting.]{Examples of ArcheType fine-tuned (top) and zero-shot (bottom) prompting.}
     \label{fig:enter-label}
    \vspace{-.8cm}
 \end{figure}

\vspace{-.15cm}
\subsection{Prompt Serialization}
\label{sec:prompt-ser}

The \textit{prompt serialization} stage transforms the context sample $S$ into a prompt format suitable for querying an LLM; this includes modification of prompts that exceed the maximum allowable length of the context window and how to reformat the table. 

\Figref{fig:enter-label} shows examples of prompts for both fine-tuned and zero-shot regimes of ArcheType. 
We style our fine-tuned prompt after the instruction-following method described in~\cite{alpaca}. We treat the semantics of the \textit{INSTRUCTION} field as a hyperparameter, and fix it at training time. The extended context includes the samples, the table name, and computed summary statistics including standard deviation, median and mode. 
In zero-shot, we again treat \textit{INSTRUCTION} as a hyperparameter, sweeping over a space of possible semantic structures. \textit{INPUT} is handled identically to fine-tuned. In zero-shot, the prompt also includes \textit{OPTIONS}, or allowable column names, from which the model is expected to choose. The suffix \textit{ANSWER:} cues the LLM to supply the label (in this case,``number'').

The heuristic optimization of this process is sometimes referred to as \emph{prompt engineering}, and is treated as an important contribution by existing zero-shot CTA methods~\cite{kayali2023chorus, korini2023column}. However, recent phenomenological studies of foundation models have raised significant doubts as to the near-term stability and long-term viability of prompt engineering as a method~\cite{sclar_quantifying_2023}.
In fine-tuned ArcheType, we fix a single prompt serialization strategy, as the prompt is learned during the fine-tuning process and has little impact on the model output, as long as it is consistent. In zero-shot ArcheType, unlike previous methods, we treat the choice of prompt as a hyperparameter. We provide experimental support for this idea in \Secref{sec:ablations-prompt-ser}.

\begin{figure}[!tb]
     \input{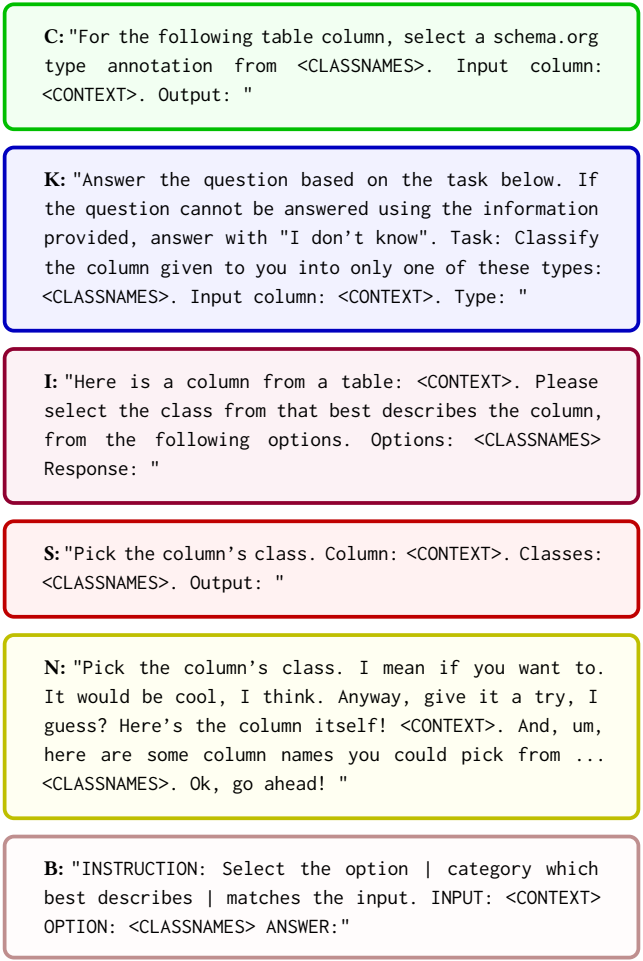}
     \vspace{-.3cm}
     \caption{\sl\textbf{Six prompt variations.} \sl In zero-shot ArcheType, we treat prompting as a hyperparameter, and sweep over six distinct prompts, each chosen according to a conceptual serialization strategy. 
     <CLASSNAMES> stands in for the label set, <CONTEXT> for the output of the context sampling step. We use two variants of the "B" prompt, with semantic differences denoted by "|".
     }
     \Description[Six Prompt Variations.]{In zero-shot ArcheType, we treat prompting as a hyperparameter, and sweep over six distinct prompts, each chosen according to a conceptual serialization strategy. 
     <CLASSNAMES> stands in for the label set, <CONTEXT> for the output of the context sampling step. We use two variants of the "B" prompt, with semantic differences denoted by "|".}
          \vspace{-.6cm}
     \label{fig:prompt-types}
 \end{figure}

\myparagraph{Serialization strategies.} We explore six distinct serialization strategies, illustrated in \Figref{fig:prompt-types}.
The strategies labeled "C" and "K" were proposed in~\cite{kayali2023chorus} and \cite{korini2023column}, respectively. The remaining serialization strategies are designed to test the effect of varying prompt length, position, and tone; "N" adopts a casual, conversational tone and uses simple language, "I" inverts the position of prompt and context, compared to the other strategies, and "S" is designed to be as short as possible while remaining clear. Our "B" prompt is written in a technical and formal tone, similar to prompt "S", but more verbose. We use a minor variant of our "B" prompt in our fine-tuned experiments; the semantic differences are shown in \cite{penfever2023archetype}.

\myparagraph{Prompt Serialization in ArcheType Zero Shot (ZS).}
We have evaluated ArcheType ZS using all six prompts in \cite{penfever2023archetype}; we report performance on the best-performing configuration. Note that we include the label set $L$ in the prompt. In order to simplify the label space further for open-source models, we attempt to detect using simple type testing whether all elements of the context are numeric; if so, we limit $L$ to labels which are numeric (selecting which labels are exclusively numeric is a one-time optimization per dataset -- on SOTAB-27, it required about five minutes).

\myparagraph{Prompt Serialization in ArcheType Fine Tuned (FT).} We follow the Alpaca instruction format described in \cite{alpaca} and omit the label set $L$ to make more efficient use of the context window.

\myparagraph{Column-at-once Serialization.} Both \cite{korini2023column} and \cite{kayali2023chorus} use \textit{table-at-once} serialization; the entire table is presented to the LLM at inference time, and all columns in that table are classified together. ArcheType uses \textit{column-at-once} serialization; only a single column to be classified is passed to the LLM. \cite{korini2023column} provides ablation studies indicating that table-at-once outperforms column-at-once on their test set, a very small subset of SOTAB. 

Table-at-once serialization, however, is impractical to implement on open-source models with small context windows, and inefficient in that it requires classification of \textit{all columns}, whether or not the classes for all columns are required.

\myparagraph{Handling Overflow.} Using the length of each prompt, we produce a conservative estimate of whether the tokenized prompt might overflow the context window. If so, we tokenize the prompt, truncate it, add the classnames and response cue to the end of the prompt, and pass it through.
Examples of serialized prompts can be found in \Figref{fig:enter-label}.

\vspace{-.25cm}
\subsection{Model Querying}
\label{sec:modelquery}

The third stage of ArcheType involves passing the serialized prompt as input to the LLM, a process which we refer to as \textit{model querying}. The key variable here is, naturally, the choice of model and, in the case of fine-tuned CTA, the approach to training said model.

\begin{algorithm}[t]
    \begin{algorithmic}[1]
\Procedure{FineTuneLM}{M,D,H}\Comment{M, an LLM, D, a fine-tuning dataset, H, hyperparameters}
    \State $\text{Tokenizer} \gets \text{LoadTokenizer}()$
    \Comment{tokenizer for the LM}
    
    \State $\text{D} \gets \text{Tokenizer}.\text{Tokenize}(\text{D})$
    \Comment{Token. the fine-tuning data}
    
    \For{$\text{epoch} = 1, \text{Hyperparameters.Epochs}$}
        \For{each $B \in D$}
            \State $\text{loss} \gets \text{M}.\text{Forward}(\text{B})$
            \Comment{Compute the forward pass}
            \State $\text{loss}.\text{backward}()$
            \Comment{Backpropagate the loss}
            \State $M \gets \text{optimizer}.\text{step}(M)$
            \Comment{Update parameters}
        \EndFor
    \EndFor
    
    \State \Return $\text{Fine-tuned Model}$
\EndProcedure
%\vspace{-.1cm}
\end{algorithmic}

    \caption{\sl\textbf{Fine-tuned ArcheType. } \sl  Fine-tuning procedure for ArcheType-LLAMA; the serialized prompts generated by ArcheType are tokenized and passed to the model in batches. The autoregressive objective during training is for the model to generate the appropriate class token, given the prompt.}
    \label{alg:fine-tune-llm}
\end{algorithm}

\myparagraph{Fine-Tuned Models.} In the fine-tuning regime, our model is a LLAMA-7B, the smallest in a  batch of LLMs from~\cite{Touvron2023LLaMAOA}. All models in the LLAMA family were pre-trained on the standard unsupervised language modeling task of next-token prediction, but had no instruction tuning as part of pre-training. In order to improve performance on instruction-following tasks, we apply the Alpaca method of~\cite{alpaca} prior to applying ArcheType. See \Algref{alg:fine-tune-llm} for an overview of the fine-tuning procedure utilized to train our model. \Figref{fig:enter-label} contains an example of a single data point in the training set.

Our results for fine-tuning are reported using a fine-tuned LLAMA-7B trained on the SOTAB-full training dataset, using our context sampling and label remapping algorithms. Following~\cite{alpaca}, we fine-tune LLAMA-7B for 3 epochs, with a learning rate of 2e-5. Fine tuning took 8-12 hours on 4x A100-80GB GPUs.

\myparagraph{Zero-Shot Models.}
In the zero-shot regime, we consider the recent open-source OPT-IML and LLAMA-2 models from~\cite{Iyer2022OPTIMLSL, touvron2023llama} as well as FLAN models introduced in~\cite{Chung2022ScalingIL, Tay2022UL2UL}. We also present results on the closed-source, private GPT family of models from OpenAI~\cite{Ouyang2022TrainingLM}. As zero-shot ArcheType is model-agnostic, we report results from the three best-performing architectures in our experiments (\Tabref{tab:main-results-zs}). 

\vspace{-.3cm}
\subsection{Label Remapping}
\label{sec:label-remap}

The fourth stage of ArcheType is \textit{label remapping}; mapping the generative output of the LLM to the space of allowed labels.
A key drawback of using standard LLMs for classification tasks (based on instruction tuning alone) is that their outputs are not guaranteed to only belong  to the provided label set. In our experiments, we found small decoder-only LLMs, such as LLAMA-7B, were particularly susceptible to this behavior.

Previous works such as~\cite{Chung2022ScalingIL} have proposed simply discarding all answers which are not an exact match for a label in the set, and measuring performance with respect to exact matches only.  Another na\"{i}ve solution is to simply map all non-matching answers to a default \texttt{null} class.

However, we find that such approaches tend to underrate what the model actually provides, particularly in the CTA context. Often, the LLM's `best guess' can be reasonably remapped to an answer in the provided label set. Formally, we frame label remapping as a function 
$REMAP(\sigma_L) : \Sigma_* \rightarrow L$.
In other words, the REMAP function is responsible for mapping arbitrary output strings (that are outputs of the LLM) to some specific label in the label set  $\sigma_L \in L$. We explore multiple approaches, described below,  and find that the optimal approach varies depending on the LLM and whether we are in a fine-tuned or zero-shot domain.

\begin{figure}[!t]
\begin{minipage}{0.48\textwidth}
    \begin{algorithm}[H]
        \input{algos/algo-remap-res}
        \caption{\sl\textbf{Remap-resample. } \sl  Remap-resample calls the LLM up to $k$ times with permuted hyperparameters in order to generate increasingly diverse responses.}
        \label{alg:remap_resample}
    \end{algorithm}
\end{minipage}
\begin{minipage}{0.48\textwidth}
    \begin{algorithm}[H]
        \input{algos/algo-remap-sim}
        \caption{\sl\textbf{Remap-similarity. } \sl  Remap-similarity maps the embedded LLM response which is not in $L$ to the embedded response in $L$ which maximizes embedding cosine similarity.}
        \label{alg:remap-similarity}
    \end{algorithm}
\end{minipage}
\Description[Remap-similarity and remap-resample.]{Remap-resample calls the LLM up to $k$ times with permuted hyperparameters in order to generate increasingly diverse responses. Remap-similarity maps the embedded LLM response which is not in $L$ to the embedded response in $L$ which maximizes embedding cosine similarity.}
\vspace{-.6cm}
\end{figure}

\emph{Remap-contains} employs the simplest strategy of checking for intersections: 
$\forall L_j \in L, (\sigma \subseteq L_j \lor L_j \subseteq \sigma) \rightarrow (\sigma_L := L_j)$.
\noindent In the case of multiple matches, we accept the longest match. This is computationally efficient but has a high rate of failure; it can therefore be used in conjunction with other label remapping strategies.

\emph{Remap-resample} (\Algref{alg:remap_resample}) utilizes the probabilistic nature of LLM outputs. We fix a hyperparameter $k$ setting both how many times we attempt the problem \textit{and} how we adjust the hyperparameters on each subsequent call. The parameter $k$ can be utilized as either an additive or a multiplicative factor; we find that additive $k$ is suitable for adjusting top\_p and repetition\_penalty, while a multiplicative factor works well for temperature. For more details on these hyperparameters, please refer to~\cite{huggingface_transformers}.

\emph{Remap-similarity} (\Algref{alg:remap-similarity}) employs a similarity-search strategy. Using an encoder-only transformer model, the input $\sigma$ is converted to a vector embedding $v_\sigma$, as are all the strings in $L$. $\forall j \in L$, we then compute the vector cosine similarity $\texttt{COSSIM}(v_\sigma, v_{L_j})$. The $\texttt{ARGMAX}$ result becomes the model's predicted class. For our experiments, we used the S3Bert model introduced in~\cite{Opitz2022SBERTSM}. This method has the advantage of always returning a solution. However, this solution may be not always the desired one; moreover,  introducing an additional model adds to overall computational complexity.

\myparagraph{Rule-Based Label Remapping.} 
We find that in many CTA datasets, certain types are straightforward to detect or correct using simple algorithmic approaches. Therefore, in order to provide a more realistic picture of how our method would perform in a real-world setting, we supplement both our baselines and ArcheType with \emph{rule-based label remapping} functions, applied both prior to and after model querying. These rules do not always lead to performance improvements, but they can save considerable time and some space in the context window; therefore, we predict they will be a valuable component of deployed CTA systems, and devote some time to studying their effects. To conserve the zero-shot nature of the problem, we limited ourselves to two hours per dataset for devising these functions. As this is a one-time cost per label set, we consider this a reasonable time budget.

In \Tabref{tab:rule-effects}, we list the number of labels for which rules led to performance improvements, and the average amount of the improvement across all models and methods. The rules lead to a moderate improvement for the different benchmarks.
\begin{table}[t]
    \caption{\sl\textbf{Manual label remapping complements LLM-CTA. } \sl  Certain labels are faster and more reliable to solve using traditional methods, rather than LLMs. We document the gains from manual label remapping on our zero-shot benchmarks. }
    \label{tab:rule-effects}
    \vspace{-.3cm}
    \resizebox{.56\columnwidth}{!}{\small \resizebox{\columnwidth}{!}{%
\begin{tabular}{@{}lrr@{}}
\toprule
\textbf{Dataset} & \multicolumn{1}{l}{\textbf{Num labels}} & \multicolumn{1}{l}{\textbf{Avg. Pct. Gain}} \\ \midrule
SOTAB            & 5                                       & 1.3\%                                        \\
D4               & 9                                       & 7.2\%                                        \\
Amstr            & 2                                       & 3.2\%                                        \\
Pubchem          & 5                                       & 9.9\%                                        \\ \bottomrule
\end{tabular}}}
    \vspace{-.3cm}
\end{table}

\myparagraph{ArcheType+.} To separate the effects of rule-based remapping from other elements of the ArcheType method, we report F1 scores with and without rule-based remapping in \Tabref{tab:main-results-ft} and \Tabref{tab:main-results-zs}. In both tables, results with rules applied are denoted with a "+" symbol.

\section{New Zero-Shot Benchmarks}
\label{sec:new-benchmarks}

Existing CTA benchmarks~\cite{Hulsebos2021GitTablesAL,TURLpaper,webtablespaper,sotab} are useful sources of real-world tabular data, but they were designed to evaluate methods that perform CTA on a fixed set of labels that belong classes in well-known ontologies and taxonomies. 

In order to probe the breadth of LLM subject knowledge and assess the effectiveness of LLM-CTA methods over rare classes with different characteristics, we create \textit{three new zero-shot column type annotation benchmarks}: D4Tables (D4-20), derived from the D4 dataset~\cite{D4paper} and \cite{nyc_opendata}, AmstrTables (Amstr-56), derived from the American Stories dataset~\cite{dell_american_2023}, and PubchemTables (Pubchem-20), derived from the Pubchem dataset~\cite{fu_pubchemrdf_2015}. Examples of our zero-shot benchmarks can be found in \Figref{fig:dataset-samples}.

Each of our benchmarks is constructed using the same general approach: we reprocess the dataset so that classes of data can be interpreted as columns, fix a random seed, and sample from the data pool to produce synthetic columns of a wide range of lengths, treating all columns as independent. This approach to CTA benchmarking stands in contrast with existing benchmarks and methods, which leverage relationships at the level of a table. However, the definition of CTA does not \emph{guarantee} the existence of such informative metadata. Furthermore, in some real-world settings, such information is not available.
We therefore regard these new benchmarks as a distinct, but valuable, way to measure progress in CTA.

We follow the approach used in \cite{sotab} and attempt to replicate, as closely as possible, the distributions encountered in \textit{real-world data}. 
This results in some column types that are extremely low-variance (such as $\texttt{ethnicity}$ in D4Tables, with only 5 unique values). In other types, the set of potential unique entries in one type is entirely subsumed by another type, e.g., \texttt{us-state, other-states} in D4-Tables. Others can be addressed model free with regex pattern matching (such as \texttt{Journal ISSN} in Pubchem). As noted in \Secref{sec:label-remap}, when such solutions are possible, we utilize them in both our baseline approaches and the ArcheType method itself.

D4, Amstr and Pubchem are generated from existing data distributions -- it is therefore possible to produce an arbitrary number of tables using them. Balancing time constraints with the desire to test a significant sample size, we heuristically select a sample size of 2000 columns, and apply this consistently to each benchmark. The complete class names for each dataset can be found in \cite{penfever2023archetype}.

\myparagraph{D4Tables.} \citet{D4paper} clustered data from NYC Open Data in an unsupervised manner, and the most coherent clusters (representing semantic types) were assigned labels; in total, 20 clusters were labeled. For more information on the clustering method and the complete label list, please refer to our repository. For our paper, we convert the clusters to columns and sample accordingly.

The classes in D4 are representative of open and public data sources, including 2 classes which correspond to city agencies, 4 classes which relate to public schools, and 5 classes which correspond to neighborhoods, streets or regions located in specific New York City Boroughs. This dataset aims to assess the model's understanding of regional information and fine-grained semantic types relevant to governments and NGOs.

\myparagraph{AmstrTables.} The American Stories dataset consists of 20 million OCR scans from the Library of Congress’s public domain Chronicling America collection. Each scan contains an article written between 1774 and 1963. We adapt this dataset for CTA by: dividing the articles in the dataset according to the state in which they were originally published; and creating additional column types for author bylines, newspaper names, and subheadings. Because this dataset was published in 2023, it is unlikely that any of the models evaluated in this study have trained on this data before, reducing concerns of potential data contamination~\cite{dell_american_2023}. Another advantage is that for the majority of column types, individual row entries are quite long, corresponding to entire newspaper articles. This phenomenon is commonplace in real-world data, but rare among academic CTA benchmarks. The classes in AmstrTables mostly pertain to journalism and history.

\myparagraph{PubchemTables.} Pubchem is the world's largest collection of freely accessible chemical information. Chemicals are identified according to their name, molecular formula, structure, biological activities, safety and toxicity information, and more. The database also contains extensive information on patents related to chemistry, such as patent 
abstracts and author names, as well as the names of scientific journals. We convert the RDF triple format provided by Pubchem to a columnar format suitable for CTA, and sample from the resulting distributions to produce our target columns. Correct classification requires specialist domain knowledge of chemistry.

\myparagraph{SOTAB-27.} 
The original SOTAB (SOTAB-91) is an unbalanced, 91-class classification problem where the task is to match each unlabeled column name with its ground-truth label. 
We created a zero-shot, simplified 27-class version of the benchmark (SOTAB-27) to reduce the semantic overlap among SOTAB labels. The tables in this dataset are identical to the original SOTAB benchmark; however, we remap the 91 labels in the full SOTAB benchmark to a smaller set of 27 labels. The exact details of the class remapping can be found in our github repository~\cite{penfever2023archetype}.

\begin{table*}[t]
    \centering
    \caption{\sl\textbf{ArcheType achieves strong performance on the SOTAB benchmark. } \sl  Without rule-based remapping, our method (ArcheType-LLAMA) achieves performance close to the best available pre-trained model (DoDuo), while requiring far less tabular pretraining data. With rule-based remapping (ArcheType-LLAMA+), our method improves upon it.}
    \label{tab:main-results-ft}
    \vspace{-.3cm}
    \begin{small}
        \begin{tabular}{l l l l l}
    \parbox{2cm}{Model Name} & \parbox{2cm}{Dataset (Train)} & \parbox{2cm}{Dataset (Eval)} & \parbox[c]{2cm}{Micro-F1} \\ 
    \midrule

    ArcheType-LLAMA+ & LLAMA + SOTAB-91  & SOTAB-91    & \textbf{85.97} $\pm 0.6$ \\
    DoDuo & VizNet + SOTAB-91 & SOTAB-91   & 84.82 $\pm 0.6$  \\
    ArcheType-LLAMA & LLAMA + SOTAB-91  & SOTAB-91    & 82.9 $\pm 0.6$ \\
    TURL & TURL-Tables + SOTAB-91 & SOTAB-91 & 78.96 $\pm 0.7$ \\
\end{tabular}
    \end{small}
                %%\vspace{-.15cm}
\end{table*}

\begin{table*}[t]
    %%\vspace{-.1cm}
\caption{\sl\textbf{ArcheType achieves state-of-the-art performance on zero-shot CTA benchmarks. } \sl  ArcheType is the among the best-performing methods across all zero-shot CTA benchmarks and model architectures in our suite. With respect to architectures, we find that neither open-source model dominates. Surprisingly, closed-source models do not dominate either; GPT wins two benchmarks, ties one and loses one. In order to ablate the effect of rule-based remapping, we separately report the performance of our models on all labels (denoted +) and on labels without rules. We also indicate the number of labels remaining in each dataset after the change. All scores are weighted Micro-F1, scale 0-100.}
\label{tab:main-results-zs}
\vspace{-.3cm}
\begin{small}
    \begin{tabular}{@{}llrrrrrrrr@{}}
    Method & Arch. & SOTAB-27+ & SOTAB-27 & D4-20+ & D4-11 & Amstr-56+ & Amstr-54 & Pubchem-20+ & Pubchem-15 \\
    \midrule 
    \multicolumn{10}{c}{\textit{OPEN-SOURCE}} \\
    ArcheType & UL2 & \textbf{60.9} $\pm 0.8$ & 58.0 $\pm 0.9$ & \textbf{82.4} $\pm 1.7$ & \textbf{70.8} $\pm 2.7$ & \textbf{35.8} $\pm 2.1$ & \textbf{32.8} $\pm 2.1$ & \textbf{70.9} $\pm 2.0$ & \textbf{61.1} $\pm 2.5$ \\ 
    C-Baseline & UL2 & 52.2 $\pm 0.8$ & 51.3 $\pm 0.9$ & 78.0 $\pm 1.8$ & \textbf{69.4} $\pm 2.7$ & 13.5 $\pm 1.5$ & 11.4 $\pm 1.4$ & 61.7 $\pm 2.1$ & 50.3 $\pm 2.5$ \\ 
    K-Baseline & UL2 & 52.8 $\pm 0.8$ & 52.5 $\pm 0.9$ & 76.7 $\pm 1.9$ & 67.6 $\pm 2.7$ & 22.4 $\pm 1.8$ & 20.1 $\pm 1.8$ & 64.8 $\pm 2.1$ & 54.7 $\pm 2.5$ \\ 
    ArcheType & T5 & \textbf{62.5} $\pm 0.8$ & \textbf{60.8} $\pm 0.9$ & \textbf{84.6} $\pm 1.6$ & \textbf{74.5} $\pm 2.6$ & 29.2 $\pm 2.0$ & 25.6 $\pm 2.0$ & \textbf{72.0} $\pm 2.0$ & \textbf{63.3} $\pm 2.4$ \\ 
    C-Baseline & T5 & 51.0 $\pm 0.8$ & 50.0 $\pm 0.9$ & 81.2 $\pm 1.7$ & \textbf{75.0} $\pm 2.6$ & 11.4 $\pm 1.4$ & 08.5 $\pm 1.3$ & \textbf{68.3} $\pm 2.0$ & \textbf{59.0} $\pm 2.5$ \\ 
    K-Baseline & T5 & 52.6 $\pm 0.8$ & 52.1 $\pm 0.9$ & 81.2 $\pm 1.7$ & \textbf{74.5} $\pm 2.6$ & 19.2 $\pm 1.7$ & 15.1 $\pm 1.6$ & 62.3 $\pm 2.1$ & 51.3 $\pm 2.5$ \\
    \midrule
    \multicolumn{10}{c}{\textit{CLOSED-SOURCE}} \\
    ArcheType & GPT & \textbf{66.0} $\pm 0.8$ & \textbf{64.3} $\pm 0.9$ & \textbf{87.3} $\pm 1.5$ & \textbf{83.0} $\pm 2.2$ & \textbf{27.2} $\pm 2.0$ & \textbf{22.5} $\pm 1.9$ & \textbf{65.9} $\pm 2.1$ & \textbf{60.2} $\pm 2.5$ \\ 
    C-Baseline & GPT & 59.3 $\pm 0.8$ & 58.5 $\pm 0.9$ & 77.7 $\pm 1.8$ & 70.8 $\pm 2.7$ & 09.0 $\pm 1.3$ & 04.9 $\pm 0.9$ & 56.0 $\pm 2.2$ & 43.0 $\pm 2.5$ \\ 
    K-Baseline & GPT & 59.3 $\pm 0.8$ & 57.2 $\pm 0.9$ & 81.8 $\pm 1.7$ & \textbf{80.9} $\pm 2.3$ & 10.0 $\pm 1.3$ & 07.9 $\pm 1.2$ & \textbf{65.8} $\pm 2.1$ & 55.8 $\pm 2.5$ \\ 
\end{tabular}

% \begin{tabular}{@{}llrrrr@{}}
% \toprule
% \textbf{Method} & \textbf{Arch.} & \multicolumn{1}{l}{\textbf{SOTAB-27}} & \multicolumn{1}{l}{\textbf{D4-20}} & \multicolumn{1}{l}{\textbf{Amstr-56}} & \multicolumn{1}{l}{\textbf{Pubchem-20}} \\ \midrule
% \multicolumn{6}{c}{\textit{PUBLIC}} \\
% ArcheType  & UL2 & 56.4 $\pm 0.9$ & \textbf{82.4} $\pm 1.8$ & \textbf{35.8} $\pm 1.9$& \textbf{70.9} $\pm 1.9$\\
% C-Baseline & UL2 & 47.3 $\pm 0.9$ & 78.0 $\pm 1.8$ & 13.5 $\pm 1.7$& 61.7 $\pm 2.0$\\
% K-Baseline & UL2 & 47.8 $\pm 0.9$ & 76.7 $\pm 1.9$ & 22.4 $\pm 1.7$ & 64.8 $\pm 2.0$\\
% ArcheType  & T5  & \textbf{58.0} $\pm 0.9$  & \textbf{84.8} $\pm 1.7$& 29.2 $\pm 1.8$  & \textbf{72.0} $\pm 1.9$ \\
% C-Baseline & T5  & 45.8 $\pm 0.8$ & 81.2 $\pm 1.8$ & 11.4 $\pm 1.7$  & \textbf{68.3} $\pm 2.0$ \\
% K-Baseline & T5  & 47.3 $\pm 0.8$ & 81.2 $\pm 1.8$ & 19.2 $\pm 1.7$  & 62.3 $\pm 2.0$ \\
% \midrule
% \multicolumn{6}{c}{\textit{PRIVATE}} \\
% ArcheType  & GPT & \textbf{63.3} $\pm 0.8$ & \textbf{87.3} $\pm 1.7$ & \textbf{27.2} $\pm 1.7$  & \textbf{71.2} $\pm 1.9$ \\
% C-Baseline & GPT & 54.3 $\pm 0.9$ & 77.7 $\pm 1.8$ & 9.0 $\pm 1.7$  & 56.0 $\pm 2.1$ \\
% K-Baseline & GPT & 54.7 $\pm 0.9$  & 81.8 $\pm 1.8$& 10.0 $\pm 1.7$  & 65.8 $\pm 2.0$ \\ \bottomrule
% \end{tabular}
\vspace{5pt}
\end{small}
\end{table*}

\begin{table*}[t]
    %%\vspace{.1cm}
\caption{\sl\textbf{ArcheType zero-shot is competitive with state-of-the-art models on well-established CTA benchmarks.} Where results were unavailable in the literature, we write n/a. 
\sl  }
\label{tab:main-results-existing}
\vspace{-.3cm}
\begin{small}
    
\begin{tabular}{@{}llllll@{}}
\textbf{Dataset} & \textbf{Metric} & \textbf{TURL-FT} & \textbf{Archetype-ZS-T5} & \textbf{Archetype-ZS-GPT4} &  \\ \midrule
T2D & Unbal. Acc. & \multicolumn{1}{r}{\textbf{96.2} $\pm 3.3$} & \multicolumn{1}{r}{\textbf{90.4} $\pm 3.4$}& \multicolumn{1}{r}{\textbf{95.8} $\pm 3.3$} &  \\
Efthymiou & Unbal. Acc. & \multicolumn{1}{r}{74.6 $\pm 3.8$} & \multicolumn{1}{r}{78.5 $\pm 3.8$} & \multicolumn{1}{r}{\textbf{95.7} $\pm 3.3$} &  \\
\bottomrule
\end{tabular}%
    \begin{tabular}{@{}llllllll@{}}
\\
\textbf{Dataset} & \textbf{Metric} & \textbf{DoDuo-VN-FT} & \textbf{DoDuo-WT-FT} & \textbf{Sherlock-FT} & \textbf{Chorus-ZS-GPT} & \textbf{Archetype-ZS-T5} & \textbf{Archetype-ZS-GPT4} \\
\midrule
T2D & Weighted F1 & \multicolumn{1}{r}{65.4 $\pm 3.9$} & \multicolumn{1}{r}{75.7 $\pm 3.8$} & \multicolumn{1}{r}{n/a} & \multicolumn{1}{r}{\textbf{92.3} $\pm 3.4$}  & \multicolumn{1}{r}{88.9 $\pm 3.4$}  & \multicolumn{1}{r}{\textbf{95.3} $\pm 3.3$} \\
VizNet-Chorus & Weighted F1 & \multicolumn{1}{r}{\textbf{90.0} $\pm 2.4$} & \multicolumn{1}{r}{81.5 $\pm 2.5$} & \multicolumn{1}{r}{\textbf{93.0 $\pm 2.4$}} & \multicolumn{1}{r}{86.5 $\pm 2.5$} & \multicolumn{1}{r}{\textbf{88.5} $\pm 2.5$} & \multicolumn{1}{r}{\textbf{90.5 $\pm 2.5$}} \\
\end{tabular}
\end{small}
\end{table*}

\vspace{-.15cm}
\section{Experiments} \label{sec:experiments}

\subsection{Experimental Setup}

\myparagraph{Fine-tuned Baselines.} 
For our fine-tuned experiments, we compare our ArcheType LLAMA-7B (\Secref{sec:modelquery}) to DoDuo~\cite{doduopaper}, the state-of-the-art model for column type annotation,  as well as TURL~\cite{TURLpaper}. 

We report DoDuo and TURL results following the approach described in \cite{sotab}, 
%who describe passing 
which passes the entire table to the model at inference time; we limit our own method to 15 samples per table. 

\myparagraph{Zero-shot Baselines.} To the best of our knowledge, there exist no \emph{open-source} CTA models that can operate in a zero-shot manner; therefore, we design strong baselines derived from zero-shot CTA methods which have been introduced specifically for use with GPT: 

\noindent \emph{C-Baseline}, based on the method in \cite{kayali2023chorus}, utilizes 
    similarity label remapping and simple random sampling, and our C-prompt.
    
\noindent \emph{K-Baseline}, derived from \cite{korini2023column}, utilizes our K-prompt, no-op label remapping and first-k-columns sampling. We omit the method described in \cite{korini2023column}, which requires a custom hash table for each problem, as this invalidates the zero-shot nature of the problem we consider here. 

For all methods, we fix 5 samples per column and provide model inputs a column-at-once manner. The prompt includes class names. 

To evaluate the robustness of the methods to variations in architecture, we evaluate each method using three different architectures: the closed-source GPT-3.5-Turbo model from OpenAI (October 2023 version) denoted GPT and GPT-4.0-Turbo model (gpt-4-turbo-preview, February 2024) denoted GPT4, 
and the open-source T5 and UL2 encoder/decoder LLMs from Google~\cite{Tay2022UL2UL}.

\myparagraph{Benchmarks.}
A variety of realistic and challenging CTA benchmarks have been developed in the last few years. Prominent among these are GitTables from~\cite{Hulsebos2021GitTablesAL}, WikiTables as modified in~\cite{TURLpaper}, and WebTables from~\cite{webtablespaper}. However, these are usually pre-processed in an ad-hoc fashion and compared against some, but not all existing methods, making it difficult to truly measure progress in the field. For this reason, we use the recent SOTAB benchmark~\cite{sotab}. SOTAB was independently tested on both state-of-the-art CTA approaches, TURL and DoDuo, making it an ideal testing ground for new CTA methods. Furthermore, it is, to the best of our knowledge, the most challenging CTA benchmark in the literature; the strongest method to date, DoDuo, achieves a Micro-F1 score of $84.8$ on SOTAB-91, while for WikiTables and VizNet it attains Micro-F1 scores between $91.47$ and $96.4$~\cite{doduopaper}.

For the zero-shot regime, we also use the benchmarks introduced in \Secref{sec:new-benchmarks}
as well as established benchmarks: T2D~\cite{chen_learning_2019},  Efthymiou~\cite{efthymiou2017matching}, and VizNet~\cite{kayali2023chorus}.

\vspace{-.4cm}
\subsection{ArcheType Effectiveness}
\label{sec:main-results}

Following~\cite{doduopaper}, we report performance using the weighted micro-F1 score--the weighted average of F1 scores based on the sample size of each class. We provide 95\% confidence intervals for all results using the normal approximation interval method. \textbf{Boldface} in tables indicates the best-performing method(s) within the error bounds.

\Tabref{tab:main-results-ft} summarizes our key results in fine-tuned CTA and \Tabref{tab:main-results-zs} shows our zero-shot findings using SOTAB and the zero-shot benchmarks (\Secref{sec:new-benchmarks}). We observe that:

1) in the fine-tuned regime, our ArcheType-LLAMA model is competitive with DoDuo, despite training on less than 1\% data; and 
2) in the zero-shot regime, ArcheType outperforms or matches baselines on all dataset/architecture pairings we evaluate. 
These results underscore the effectiveness of ArcheType and serve as evidence that, LLMs can enable CTA methods that are not just robust to distribution shift, but that handle open-label sets defined at inference time, including rare types.

We also compare our zero-shot ArcheType to prior CTA approaches on established benchmarks, specifically: TURL, fine-tuned on the T2D \cite{chen_learning_2019} and Efthymiou \cite{efthymiou2017matching} benchmarks; CHORUS~\cite{kayali2023chorus}, zero-shot on T2D and a stratified sample of the VizNet dataset (VizNet-CHORUS); DoDuo, fine-tuned on VizNet (VN) and WikiTables (WT) and evaluated on VizNet-CHORUS; and Sherlock, fine-tuned on VizNet and evaluated on VizNet-CHORUS. In all cases, we follow as closely as possible the methodology of the aforementioned authors, adopting their metrics.

As \Tabref{tab:main-results-existing} shows, ArcheType's performance is comparable to that of the other systems (both fine-tuned and zero-shot) even when using the smallest (T5) backbone.

\vspace{-.3cm}
\subsection{Observations}
\label{sec:observations}

A detailed analysis of our results has both confirmed our hypotheses regarding LLMs as well as uncovered insights into some of their limitations. We summarize these below.

\myparagraphemph{LLMs contain sufficient world knowledge to perform zero-shot CTA on domain-specific classes.} We find that LLM performance is consistently strong across datasets and across benchmarks, emphasizing the generality of LLM-CTA, compared to fine-tuned methods such as DoDuo. In PubchemTables, we observe that models are consistently able to disambiguate challenging classes such as \textit{disease, chemical, taxonomy, patent, SMILES (simplified molecular input line entry system)}, and \textit{molecular formula}. On D4Tables, they are able to disambiguate the names of NYC public schools and NYC governmental agencies, as well as identify locations. With $\phi = 5$, we find that ArcheType-T5 and UL2 are able to correctly identify whether the addresses are in Queens, the Bronx, Brooklyn or Manhattan more than 50\% of the time, on average. ArcheType-GPT is even more impressive; it is able to accurately classify regions in all five boroughs more than 87\% of the time, on average. Class-specific accuracies for our zero-shot models can be found in \Tabref{tab:d4-class}, \Tabref{tab:sotab-class} and \Tabref{tab:pubchem-class}. 

\myparagraphemph{Model error tends to be patterned and predictable when the prompt space is fixed. } When zero-shot CTA fails, it tends to do so in ways that are patterned and predictable, making it easier to correct errors. The most common failure mode is class bias in favor of certain dataset classes over others. For any given prompt/model/dataset triple, this results in certain columns with near-perfect accuracy and others with near-zero accuracy, with the confusion matrix heavily concentrated in a few classes. We provide examples of this phenomenon in \Tabref{tab:d4-class}, \Tabref{tab:sotab-class} and \Tabref{tab:pubchem-class}. The confusion matrices for our zero-shot models can be found in our \href{https://arxiv.org/pdf/2310.18208}{GitHub repository}.

\myparagraphemph{Simple factors can be used to estimate zero-shot CTA performance.} Zero-shot performance is stronger on datasets such as PubchemTables and D4Tables; we attribute this to smaller label spaces, smaller individual sample sizes, and a high degree of intra-column similarity and a low degree of inter-column similarity. Amstr, which has more than twice as many labels as the next-largest dataset and a high degree of inter-column similarity (because the vast majority of the labels in the dataset correspond to newspaper articles drawn from the same general distribution), is the most challenging dataset in our benchmark.

\myparagraphemph{ArcheType using open-source models is highly competitive with closed-source models.} ArcheType CTA works well with a range of LLMs, small and large, open-source and closed-source, indicating that CTA benefits from flexibility in the model querying phase. Although GPT tends to have the strongest performance, the difference is not very large, and on PubChem and Amstr, GPT underperforms compared to the open-source models.

\subsection{Ablation Studies}

\subsubsection{Ablations on Context Sampling}
\label{sec:ablations-context-sampling}
\begin{figure}[t]
    \resizebox{.7\columnwidth}{!}{
    \begin{tikzpicture}
\definecolor{orange}{RGB}{255,140,0}
\definecolor{navyblue}{RGB}{0,0,128}
\definecolor{forestgreen}{RGB}{34,139,34}
\definecolor{lightgray204}{RGB}{204,204,204}
\begin{axis}
[
ybar,        
width=8cm,        
height=5cm,        
ymin=40,        
ymax=65,        
xlabel=Sampling method,
xlabel style={anchor=north,font=\normalsize},
ylabel=Micro-F1,        
xticklabel style={rotate=0,font=\small},   
xtick=data,      
xticklabels={SRS, FS, ArcheType},    
bar width=0.4cm,        
legend style={at={(0.75,1.35)},
fill opacity=0.8,
draw opacity=1,
text opacity=1,
draw=lightgray204,
anchor=north},        
enlarge x limits=0.2,        
]

\addlegendentry{ZS-T5}
\addplot[fill=orange] 
coordinates{
    (0,53.2) (1,53.3) (2,57.5)
};
\addlegendentry{ZS-UL2}
\addplot[fill=navyblue] 
coordinates{
    (0,45.6) (1,45.4) (2,54.5)
};
\addlegendentry{ZS-GPT}
\addplot[fill=forestgreen] 
coordinates{
    (0,58.9) (1,58.7) (2,63.3)
};
\end{axis}
\end{tikzpicture}}
    \vspace{-.3cm}
    \caption{\sl\textbf{ArcheType sampling outperforms baseline methods.} The sampling method used by Zero-shot ArcheType using different architectures (GPT, UL2, and T5) on the SOTAB-27 dataset, substantially outperforms simple random sampling (SRS) and first-k-entries sampling (FS), as used in \cite{kayali2023chorus, korini2023column}.}
    \Description[ArcheType sampling outperforms baseline methods.]{ArcheType sampling outperforms baseline methods. The sampling method used by Zero-shot ArcheType using different architectures (GPT, UL2, and T5) on the SOTAB-27 dataset, substantially outperforms simple random sampling (SRS) and first-k-entries sampling (FS), as used in Chorus and Korini.}\label{fig:sampling_strategy_ablations}        
%%\vspace{-.35cm}
\end{figure}
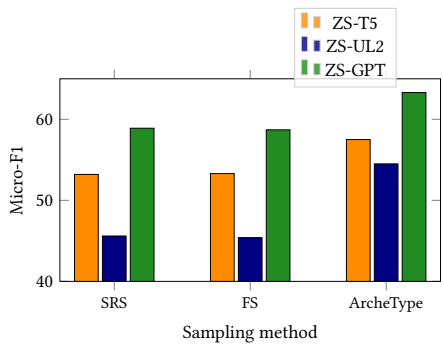

In \Figref{fig:sampling_strategy_ablations}, we ablate our choice of strategy using the SOTAB dataset, and find that ArcheType sampling consistently outperforms baseline methods. 

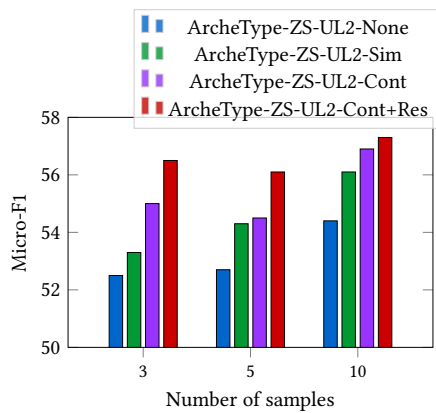
\begin{figure}
%    \begin{minipage}{0.65\columnwidth}
    \resizebox{.7\columnwidth}{!}{
    \begin{tikzpicture}
\definecolor{myblue}{RGB}{0, 102, 204}
\definecolor{mygreen}{RGB}{0, 153, 51}
\definecolor{myorange}{RGB}{255, 102, 0}
\definecolor{mypurple}{RGB}{153, 51, 255}
\definecolor{myred}{RGB}{204, 0, 0}
\definecolor{lightgray204}{RGB}{204,204,204}

\begin{axis}[
    ybar,
    width=7cm,
    height=5cm,
    ymin=50,
    ymax=58,
    xlabel=Number of samples,
    xlabel style={anchor=north, font=\normalsize},
    ylabel=Micro-F1,
    xticklabel style={rotate=0, font=\small},
    xtick=data,
    xticklabels={3, 5, 10},
    bar width=0.2cm,
    legend style={
        at={(0.6,1.47)},
        fill opacity=0.8,
        draw opacity=1,
        text opacity=1,
        draw=lightgray204,
        anchor=north,
    },
    enlarge x limits=0.35,
]

\addlegendentry{ArcheType-ZS-UL2-None}
\addplot[fill=myblue]
coordinates {
    (0,52.5) (1,52.7) (2,54.4)
};
\addlegendentry{ArcheType-ZS-UL2-Sim}
\addplot[fill=mygreen]
coordinates {
    (0,53.3) (1,54.3) (2,56.1)
};
\addlegendentry{ArcheType-ZS-UL2-Cont}
\addplot[fill=mypurple]
coordinates {
    (0,55.0) (1,54.5) (2,56.9)
};
\addlegendentry{ArcheType-ZS-UL2-Cont+Res}
\addplot[fill=myred]
coordinates {
    (0,56.5) (1,56.1) (2,57.3)
};
\end{axis}

\end{tikzpicture}}
    \vspace{-.3cm}
    \caption{\sl\textbf{ArcheType performance is affected by context size and label remapping.} The model benefits from increasing the context size from 3 to 10 samples. All methods outperform a baseline no-op method. CONTAINS+RESAMPLE performs best at every context scale.}
    \Description[ArcheType performance is affected by context size and label remapping.]{ArcheType performance is affected by context size and label remapping. The model benefits from increasing the context size from 3 to 10 samples. All methods outperform a baseline no-op method. CONTAINS+RESAMPLE performs best at every context scale.}\label{fig:algo_ablations_zs}
%    \end{minipage}
\vspace{-.4cm}
\end{figure}

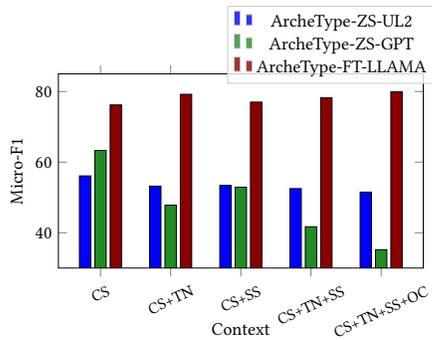
\begin{figure}[t]
     \resizebox{.7\columnwidth}{!}{
        \begin{tikzpicture}
\definecolor{darkgray176}{RGB}{176,176,176}
\definecolor{darkred}{RGB}{139,0,0}
\definecolor{forestgreen}{RGB}{34,139,34}
\definecolor{lightgray204}{RGB}{204,204,204}

\begin{axis}
[
ybar,
width=8cm,
height=5cm,
ymin=30,
ymax=85,
xlabel=Context,
xlabel style={at={(0.5,-0.25)},anchor=north, font=\normalsize},
ylabel=Micro-F1,
xticklabel style={rotate=25, font=\small},
xtick=data,
xticklabels={CS,CS+TN,CS+SS,CS+TN+SS,CS+TN+SS+OC},
bar width=0.2cm,
legend style={at={(0.75,1.35)},
fill opacity=0.8,
draw opacity=1,
text opacity=1,
draw=lightgray204,
anchor=north},
enlarge x limits=0.15,
]

\addlegendentry{ArcheType-ZS-UL2}
\addplot[fill=blue] 
coordinates{
    (0,56.1) (1,53.2) (2,53.4) (3,52.5) (4,51.5)
};

\addlegendentry{ArcheType-ZS-GPT}
\addplot[fill=forestgreen] 
coordinates{
    (0,63.3) (1,47.8) (2,52.9) (3,41.7) (4,35.2)
};

\addlegendentry{ArcheType-FT-LLAMA}
\addplot[fill=darkred] 
coordinates{
    (0,76.2) (1,79.21) (2,77.0) (3,78.2) (4,79.9)
};

\end{axis}
\end{tikzpicture}
        }
        \vspace{-.3cm}
        \caption{\sl\textbf{Expanding feature selection during context sampling improves fine-tuned CTA performance, but degrades zero-shot performance.} A fine-tuned ArcheType-LLAMA model is able to learn helpful associations from features such as summary statistics (SS), table filenames (TN), and other columns (OC), but that same information is not helpful when serialized in a zero-shot prompt, even when the prompt is customized to explain what each feature is. 
        }
        \Description[Expanding feature selection during context sampling improves fine-tuned CTA performance, but degrades zero-shot performance.]{Expanding feature selection during context sampling improves fine-tuned CTA performance, but degrades zero-shot performance. A fine-tuned ArcheType-LLAMA model is able to learn helpful associations from features such as summary statistics (SS), table filenames (TN), and other columns (OC), but that same information is not helpful when serialized in a zero-shot prompt, even when the prompt is customized to explain what each feature is.}\label{fig:context_type_ablations}        
%    \end{minipage}
\end{figure}

\myparagraphemph{Sample size.} The sample size $0 < \phi \leq c$ is a hyperparameter fixed at training time (in the case of fine-tuned) or inference time (in the case of zero-shot).  
In general, we observe in \Figref{fig:algo_ablations_zs} that larger values of $\phi$ tend to result in better model performance, with the trade-off of slower inference and a larger number of truncated prompts.

\myparagraphemph{Feature selection.} In \Figref{fig:context_type_ablations}, we ablate our feature selection method, and find that ArcheType-FT benefits from each feature added, and ArcheType-ZS exhibits the opposite trend, even when we clearly identify the different types of incoming context: 

\begin{small}
\begin{verbatim}
    TABLE NAME: " sourced from the 
    table named " + <TABLE_NAME>
    OTHER COLUMNS: "For additional 
    context, here are some entries 
    from other columns in the table:
    " + <OTHER_COLUMNS>
\end{verbatim}
\end{small}

\noindent We consider the effective use of additional features an important area for future zero-shot CTA research.

\subsubsection{Ablations on Prompt Serialization}
\label{sec:ablations-prompt-ser}

We observe that improvements based on prompt serialization are quite sensitive to small changes in prompts; furthermore, the effects of these small changes differ depending on the LLM used.
In \Tabref{tab:prompt-ablation}, 
We explore six different prompts, labeled C(horus-style), K(orini-style), I(nverted), S(hort), N(oisy), B(aseline) (\Secref{sec:prompt-ser}). The first two prompt styles are adapted from \cite{kayali2023chorus, korini2023column}, respectively.  We provide examples of each prompt in \Figref{fig:prompt-types}. 
We test these prompts on SOTAB-27, holding other factors constant,  across three architectures. 
As \Tabref{tab:prompt-ablation} shows: (1) All models are very sensitive to the choice of prompt; and (2) No prompt is a top-two performer on all three models.
This supports our choice of using prompt serialization strategy as a hyperparameter. 
We also experimented with changing the label associated with a class and the \textit{position} of a label in the string, and observed that these can have unpredictable effects on performance; namely, performance of relabeled class may not change, while performances of classes with the \textit{same} labels \textit{does} change. See App. ~\Tabref{tab:cn-ablation} for details.

\begin{table}[t]
\begin{small}
    \begin{tabular}{@{}llrrr@{}}
\textbf{Prompt} & \multicolumn{1}{l}{\textbf{T5}} & \multicolumn{1}{l}{\textbf{GPT}} & \multicolumn{1}{l}{\textbf{UL2}} \\
\cellcolor[HTML]{F2FFF2}{C}   & 49.4   & 57.6   & \cellcolor[HTML]{00FF00}56.4    \\
\cellcolor[HTML]{F2F2FF}{K}   & \cellcolor[HTML]{00FF00}54.0    & \cellcolor[HTML]{EA9999}53.2    & 53.4 \\
\cellcolor[HTML]{F8F2F8}{I}   & 52.1  & 62.5 & \cellcolor[HTML]{FFFF00}55.2   \\
\cellcolor[HTML]{FFF2F2}{S}   & \cellcolor[HTML]{FFFF00}53.0    & \cellcolor[HTML]{00FF00}64.6    & 54.5 \\
\cellcolor[HTML]{FFFFF2}{N}   & 48.6 & 61.4 & \cellcolor[HTML]{EA9999}47.2    \\
\cellcolor[HTML]{FFFBFB}{B}   & \cellcolor[HTML]{EA9999}47.1   & \cellcolor[HTML]{FFFF00}63.4    & 52.1
\end{tabular}
\end{small}
\caption{\sl\textbf{Prompt serialization has unpredictable effects across models.} \sl A particular prompt can be engineered to perform well on a given model and fail to reproduce on others. Results shown are zero-shot Micro-F1 scores on the SOTAB-27 dataset. The best-performing prompt is highlighted in green, the second-best in yellow, and the lowest-performing in red. %Table best viewed in color.
}
\label{tab:prompt-ablation}
\vspace{-.5cm}
\end{table}

\myparagraphemph{Prompt serialization as a hyperparameter.}
Our method treats prompt serialization and classname selection as \textit{tunable hyperparameters} to be optimized and reported alongside experimental results.
With the understanding that \textit{any reasonable prompt is as likely to succeed as any other}~\cite{sclar_quantifying_2023}, for each model-dataset pair, we conduct a grid search over our six prompt styles, each of which is stylistically distinct but similar in content and meaning. All prompts follow general best practices as described in \cite{Touvron2023LLaMAOA}, using capital letters, colons and line breaks to delineate instructions, label sets and context, but otherwise vary widely.

\subsubsection{Ablations on Model Querying}
\label{sec:ablations-model-querying}

The space of both open and closed LLMs has exploded of late, and the performance of these models on benchmarks can vary considerably. Rather than attempt an exhaustive comparison which would quickly grow out-of-date, we select strong representative models to stand for different categories of LLM which are frequently encountered in the literature. We find that \textit{parameter count is not predictive of CTA performance}, and that \textit{encoder-decoder architectures outperform decoder-only architectures} on this task. Due to space limitations, we include further details and experimental support in \href{https://arxiv.org/abs/2310.18208}{\cite{archetype-arxiv}}.

\subsubsection{Ablations on Label Remapping}
\label{sec:ablations-label-remapping}

The choice of label remapping algorithm can substantially impact model performance; however, the number of remapped labels depends considerably on the selections made in the other three elements of the LLM-CTA method, as well as the dataset itself.
We found a positive correlation between the number of remapped labels and model accuracy. 

As \Figref{fig:algo_ablations_zs} shows,  CONTAINS+RESAMPLE (Cont+Res) outperforms the other remapping strategies  for all sample sizes. 

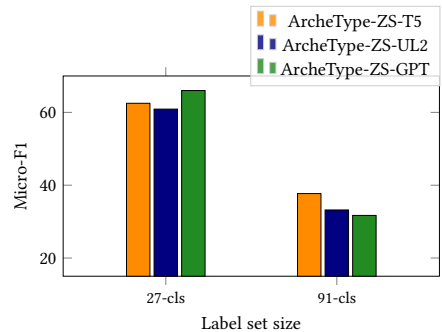
\begin{figure}[t!]
      \resizebox{.7\columnwidth}{!}{
        \begin{tikzpicture}
\definecolor{orange}{RGB}{255,140,0}
\definecolor{navyblue}{RGB}{0,0,128}
\definecolor{forestgreen}{RGB}{34,139,34}
\definecolor{lightgray204}{RGB}{204,204,204}
\begin{axis}
[
ybar,        
width=8cm,        
height=5cm,        
ymin=15,        
ymax=70,        
xlabel=Label set size,
xlabel style={anchor=north,font=\normalsize},
ylabel=Micro-F1,        
xticklabel style={rotate=0,font=\small},   
xtick=data,      
xticklabels={27-cls, 91-cls},    
bar width=0.4cm,        
legend style={at={(0.75,1.35)},
fill opacity=0.8,
draw opacity=1,
text opacity=1,
draw=lightgray204,
anchor=north},        
enlarge x limits=0.6,        
]

\addlegendentry{ArcheType-ZS-T5}
\addplot[fill=orange] 
coordinates{
    (0,62.5) (1,37.7)
};
\addlegendentry{ArcheType-ZS-UL2}
\addplot[fill=navyblue] 
coordinates{
    (0,60.9) (1,33.2)
};
\addlegendentry{ArcheType-ZS-GPT}
\addplot[fill=forestgreen] 
coordinates{
    (0,66.0) (1,31.7)
};
\end{axis}
\end{tikzpicture}
       }
        \vspace{-.4cm}
        \caption{\sl\textbf{Zero-shot performance degrades with large label sets.} Both open and closed-source LLMs for zero-shot CTA  struggle when the size of the label set grows large, compared to fine-tuned CTA.}
        \Description[Zero-shot performance degrades with large label sets.]{Zero-shot performance degrades with large label sets. Both open and closed-source LLMs for zero-shot CTA currently struggle when the size of the label set grows large, compared to fine-tuned CTA.}\label{fig:label_set_size}  
    \vspace{-.4cm}
\end{figure}

\subsection{Limitations}

Like~\cite{Narayan2022CanFM} and~\cite{Hegselmann2022TabLLMFC}, we find that there is good reason to be optimistic about the potential for large language models to dramatically impact CTA and downstream data integration and discovery applications. Despite their strong performance, we note some limitations.

\myparagraphemph{Context window lengths.} The ArcheType-LLAMA method requires only 15 samples per column to reach parity with DoDuo, but it is difficult to exceed 15 samples without truncating individual examples. For that same reason, it is difficult to present large numbers of classes to zero-shot models. This limitation may be short-lived, as context windows are already reaching 200k tokens~\cite{anthropic2024}.

\myparagraphemph{High parameter counts.} Despite generalizing very well to distribution shifts, ArcheType models have very high parameter counts when compared to previous deep learning solutions. We find that increased parameter counts are likely necessary in order for the model to contain sufficient world knowledge to be applicable for CTA ``in-the-wild''; however, the value added via zero-shot CTA methods will have to be weighed against their higher latency, energy, and carbon costs when they are deployed.

\myparagraphemph{Context sampling.} As noted in \Secref{sec:ablations-context-sampling}, zero-shot ArcheType models struggle when new features are added during context sampling. We consider this an important area of future work.

\myparagraphemph{Numeric attributes.} Although we benchmark ArcheType on all data types, we see the system as being primarily useful for semantic types (categorical or textual columns).  Simpler approaches are likely work just as well (or perhaps even better) for purely numeric or alphanumeric columns.

\myparagraphemph{Label set size.} As \Figref{fig:label_set_size} shows, all model architectures studied in this paper struggle to maintain their performance as the label set grows large, even when the context window is not exceeded. A possible reason for this is the difficulty in disambiguating several similar semantic concepts given only a brief label.

\section{Conclusions and Future Work}

We introduce ArcheType, a novel CTA approach centered around LLMs. We show that with effective context sampling and label remapping, (a) LLMs can be made highly competitive with SOTA CTA models in the fine-tuned setting, and (b) LLMs are both easier to apply, and more accurate than existing deep models in the zero-shot domain. Using newly curated benchmarks (\Secref{sec:new-benchmarks}), we show that LLM-based CTA can generalize to considerable distribution shifts, making them ideally suited for real-world tasks.

We anticipate that methods building upon ArcheType can be useful in a variety of downstream dataset creation, curation, and processing tasks. In the future, we will explore the possibility of extending our methods to novel data tasks, such as semantic joinability, column property annotation, and dataset synthesis.

\begin{acks}
This work was supported by NSF awards IIS-2106888, CMMI-2146306, and CCF-2046235; the AI Research Institutes program supported by NSF and USDA-NIFA under Award No. 2021-67021-35329; and DARPA D3M and ASKEM programs. Opinions, findings, conclusions, or recommendations expressed in this material are those of the authors and do not reflect the views of NSF, USDA, or DARPA.
\end{acks}

\bibliographystyle{ACM-Reference-Format}
\bibliography{main}


\begin{thebibliography}{54}

%%% the Web address.

\ifx \showCODEN    \undefined \def \showCODEN     #1{\unskip}     \fi
\ifx \showDOI      \undefined \def \showDOI       #1{#1}\fi
\ifx \showISBNx    \undefined \def \showISBNx     #1{\unskip}     \fi
\ifx \showISBNxiii \undefined \def \showISBNxiii  #1{\unskip}     \fi
\ifx \showISSN     \undefined \def \showISSN      #1{\unskip}     \fi
\ifx \showLCCN     \undefined \def \showLCCN      #1{\unskip}     \fi
\ifx \shownote     \undefined \def \shownote      #1{#1}          \fi
\ifx \showarticletitle \undefined \def \showarticletitle #1{#1}   \fi
\ifx \showURL      \undefined \def \showURL       {\relax}        \fi
% invisible to TeX
\providecommand\bibfield[2]{#2}
\providecommand\bibinfo[2]{#2}
\providecommand\natexlab[1]{#1}
\providecommand\showeprint[2][]{arXiv:#2}

\bibitem[Anthropic(2024)]%
        {anthropic2024}
\bibfield{author}{\bibinfo{person}{Anthropic}.}
  \bibinfo{year}{2024}\natexlab{}.
\newblock \bibinfo{title}{Introducing the next generation of Claude}.
\newblock
\newblock


\bibitem[Auer et~al\mbox{.}(2007)]%
        {auer2007dbpedia}
\bibfield{author}{\bibinfo{person}{S{\"o}ren Auer}, \bibinfo{person}{Christian
  Bizer}, \bibinfo{person}{Georgi Kobilarov}, \bibinfo{person}{Jens Lehmann},
  \bibinfo{person}{Richard Cyganiak}, {and} \bibinfo{person}{Zachary Ives}.}
  \bibinfo{year}{2007}\natexlab{}.
\newblock \showarticletitle{Dbpedia: A nucleus for a web of open data}. In
  \bibinfo{booktitle}{\emph{international semantic web conference}}.
  \bibinfo{publisher}{Springer}, \bibinfo{pages}{722--735}.
\newblock


\bibitem[Bommasani et~al\mbox{.}(2021)]%
        {bommasani2021opportunities}
\bibfield{author}{\bibinfo{person}{Rishi Bommasani}, \bibinfo{person}{Drew~A
  Hudson}, \bibinfo{person}{Ehsan Adeli}, \bibinfo{person}{Russ Altman},
  \bibinfo{person}{Simran Arora}, \bibinfo{person}{Sydney von Arx},
  \bibinfo{person}{Michael~S Bernstein}, \bibinfo{person}{Jeannette Bohg},
  \bibinfo{person}{Antoine Bosselut}, \bibinfo{person}{Emma Brunskill},
  {et~al\mbox{.}}} \bibinfo{year}{2021}\natexlab{}.
\newblock \showarticletitle{On the opportunities and risks of foundation
  models}.
\newblock \bibinfo{journal}{\emph{arXiv preprint arXiv:2108.07258}}
  (\bibinfo{year}{2021}).
\newblock


\bibitem[Bommasani et~al\mbox{.}(2023)]%
        {Liang2022HolisticEO}
\bibfield{author}{\bibinfo{person}{Rishi Bommasani}, \bibinfo{person}{Percy
  Liang}, {and} \bibinfo{person}{Tony Lee}.} \bibinfo{year}{2023}\natexlab{}.
\newblock \showarticletitle{Holistic evaluation of language models}.
\newblock \bibinfo{journal}{\emph{Annals of the New York Academy of Sciences}}
  \bibinfo{volume}{1525}, \bibinfo{number}{1} (\bibinfo{year}{2023}),
  \bibinfo{pages}{140--146}.
\newblock


\bibitem[Cafarella et~al\mbox{.}(2008)]%
        {webtablespaper}
\bibfield{author}{\bibinfo{person}{Michael~J. Cafarella}, \bibinfo{person}{Alon
  Halevy}, \bibinfo{person}{Daisy~Zhe Wang}, \bibinfo{person}{Eugene Wu}, {and}
  \bibinfo{person}{Yang Zhang}.} \bibinfo{year}{2008}\natexlab{}.
\newblock \showarticletitle{WebTables: Exploring the Power of Tables on the
  Web}.
\newblock \bibinfo{journal}{\emph{Proc. VLDB Endow.}} \bibinfo{volume}{1},
  \bibinfo{number}{1} (\bibinfo{date}{aug} \bibinfo{year}{2008}),
  \bibinfo{pages}{538–549}.
\newblock
\showISSN{2150-8097}


\bibitem[Chen et~al\mbox{.}(2019)]%
        {chen_learning_2019}
\bibfield{author}{\bibinfo{person}{Jiaoyan Chen}, \bibinfo{person}{Ernesto
  Jimenez-Ruiz}, \bibinfo{person}{Ian Horrocks}, {and} \bibinfo{person}{Charles
  Sutton}.} \bibinfo{year}{2019}\natexlab{}.
\newblock \bibinfo{title}{Learning {Semantic} {Annotations} for {Tabular}
  {Data}}.
\newblock
\newblock
\urldef\tempurl%
\url{http://arxiv.org/abs/1906.00781}
\showURL{%
\tempurl}
\newblock
\shownote{arXiv:1906.00781 [cs]}.


\bibitem[Chen et~al\mbox{.}(2024)]%
        {chen2023analyzing}
\bibfield{author}{\bibinfo{person}{Lingjiao Chen}, \bibinfo{person}{Matei
  Zaharia}, {and} \bibinfo{person}{James Zou}.}
  \bibinfo{year}{2024}\natexlab{}.
\newblock \showarticletitle{{How} {Is} {ChatGPT}\textquoteright{}s {Behavior}
  {Changing} {Over} {Time}?}
\newblock \bibinfo{journal}{\emph{Harvard Data Science Review}}
  \bibinfo{volume}{6}, \bibinfo{number}{2} (\bibinfo{date}{mar 12}
  \bibinfo{year}{2024}).
\newblock
\newblock
\shownote{https://hdsr.mitpress.mit.edu/pub/y95zitmz}.


\bibitem[Chung et~al\mbox{.}(2022)]%
        {Chung2022ScalingIL}
\bibfield{author}{\bibinfo{person}{Hyung~Won Chung}, \bibinfo{person}{Le Hou},
  \bibinfo{person}{Shayne Longpre}, \bibinfo{person}{Barret Zoph},
  \bibinfo{person}{Yi Tay}, \bibinfo{person}{William Fedus},
  \bibinfo{person}{Eric Li}, \bibinfo{person}{Xuezhi Wang},
  \bibinfo{person}{Mostafa Dehghani}, \bibinfo{person}{Siddhartha Brahma},
  \bibinfo{person}{Albert Webson}, \bibinfo{person}{Shixiang~Shane Gu},
  \bibinfo{person}{Zhuyun Dai}, \bibinfo{person}{Mirac Suzgun},
  \bibinfo{person}{Xinyun Chen}, \bibinfo{person}{Aakanksha Chowdhery},
  \bibinfo{person}{Sharan Narang}, \bibinfo{person}{Gaurav Mishra},
  \bibinfo{person}{Adams Yu}, \bibinfo{person}{Vincent~Y. Zhao},
  \bibinfo{person}{Yanping Huang}, \bibinfo{person}{Andrew~M. Dai},
  \bibinfo{person}{Hongkun Yu}, \bibinfo{person}{Slav Petrov},
  \bibinfo{person}{Ed~H. Chi}, \bibinfo{person}{Jeff Dean},
  \bibinfo{person}{Jacob Devlin}, \bibinfo{person}{Adam Roberts},
  \bibinfo{person}{Denny Zhou}, \bibinfo{person}{Quoc~V. Le}, {and}
  \bibinfo{person}{Jason Wei}.} \bibinfo{year}{2022}\natexlab{}.
\newblock \showarticletitle{Scaling Instruction-Finetuned Language Models}.
\newblock \bibinfo{journal}{\emph{CoRR}}  \bibinfo{volume}{abs/2210.11416}
  (\bibinfo{year}{2022}).
\newblock
\showeprint{2210.11416}


\bibitem[Dell et~al\mbox{.}(2024)]%
        {dell_american_2023}
\bibfield{author}{\bibinfo{person}{Melissa Dell}, \bibinfo{person}{Jacob
  Carlson}, \bibinfo{person}{Tom Bryan}, \bibinfo{person}{Emily Silcock},
  \bibinfo{person}{Abhishek Arora}, \bibinfo{person}{Zejiang Shen},
  \bibinfo{person}{Luca D'Amico-Wong}, \bibinfo{person}{Quan Le},
  \bibinfo{person}{Pablo Querubin}, {and} \bibinfo{person}{Leander Heldring}.}
  \bibinfo{year}{2024}\natexlab{}.
\newblock \showarticletitle{American stories: A large-scale structured text
  dataset of historical us newspapers}.
\newblock \bibinfo{journal}{\emph{Advances in Neural Information Processing
  Systems}}  \bibinfo{volume}{36} (\bibinfo{year}{2024}).
\newblock


\bibitem[Deng et~al\mbox{.}(2022)]%
        {TURLpaper}
\bibfield{author}{\bibinfo{person}{Xiang Deng}, \bibinfo{person}{Huan Sun},
  \bibinfo{person}{Alyssa Lees}, \bibinfo{person}{You Wu}, {and}
  \bibinfo{person}{Cong Yu}.} \bibinfo{year}{2022}\natexlab{}.
\newblock \showarticletitle{{TURL}: {Table} {Understanding} through
  {Representation} {Learning}}.
\newblock \bibinfo{journal}{\emph{SIGMOD Rec. Association for Computing
  Machinery}} \bibinfo{volume}{51}, \bibinfo{number}{1} (\bibinfo{date}{June}
  \bibinfo{year}{2022}), \bibinfo{pages}{33--40}.
\newblock
\showISSN{0163-5808}


\bibitem[Efthymiou et~al\mbox{.}(2017)]%
        {efthymiou2017matching}
\bibfield{author}{\bibinfo{person}{Vasilis Efthymiou}, \bibinfo{person}{Oktie
  Hassanzadeh}, \bibinfo{person}{Mariano Rodriguez-Muro}, {and}
  \bibinfo{person}{Vassilis Christophides}.} \bibinfo{year}{2017}\natexlab{}.
\newblock \showarticletitle{Matching web tables with knowledge base entities:
  from entity lookups to entity embeddings}. In
  \bibinfo{booktitle}{\emph{International Semantic Web Conference}}. Springer,
  \bibinfo{pages}{260--277}.
\newblock


\bibitem[Feuer and Liu(2023)]%
        {penfever2023archetype}
\bibfield{author}{\bibinfo{person}{Benjamin Feuer} {and}
  \bibinfo{person}{Yurong Liu}.} \bibinfo{year}{2023}\natexlab{}.
\newblock \bibinfo{title}{The ArcheType System}.
\newblock \bibinfo{howpublished}{\url{https://github.com/penfever/ArcheType}}.
\newblock


\bibitem[Feuer et~al\mbox{.}(2023)]%
        {archetype-arxiv}
\bibfield{author}{\bibinfo{person}{Benjamin Feuer}, \bibinfo{person}{Yurong
  Liu}, \bibinfo{person}{Chinmay Hegde}, {and} \bibinfo{person}{Juliana
  Freire}.} \bibinfo{year}{2023}\natexlab{}.
\newblock \showarticletitle{ArcheType: A Novel Framework for Open-Source Column
  Type Annotation using Large Language Models}.
\newblock \bibinfo{journal}{\emph{arXiv preprint arXiv:2310.18208}}
  (\bibinfo{year}{2023}).
\newblock


\bibitem[Fu et~al\mbox{.}(2015)]%
        {fu_pubchemrdf_2015}
\bibfield{author}{\bibinfo{person}{Gang Fu}, \bibinfo{person}{Colin Batchelor},
  \bibinfo{person}{Michel Dumontier}, \bibinfo{person}{Janna Hastings},
  \bibinfo{person}{Egon Willighagen}, {and} \bibinfo{person}{Evan Bolton}.}
  \bibinfo{year}{2015}\natexlab{}.
\newblock \showarticletitle{{PubChemRDF}: towards the semantic annotation of
  {PubChem} compound and substance databases}.
\newblock \bibinfo{journal}{\emph{Journal of Cheminformatics}}
  \bibinfo{volume}{7} (\bibinfo{date}{July} \bibinfo{year}{2015}),
  \bibinfo{pages}{34}.
\newblock
\showISSN{1758-2946}


\bibitem[Gibbons(2016)]%
        {distinctvalues}
\bibfield{author}{\bibinfo{person}{Phillip~B Gibbons}.}
  \bibinfo{year}{2016}\natexlab{}.
\newblock \showarticletitle{Distinct-values estimation over data streams}.
\newblock In \bibinfo{booktitle}{\emph{Data Stream Management: Processing
  High-Speed Data Streams}}. \bibinfo{publisher}{Springer},
  \bibinfo{pages}{121--147}.
\newblock


\bibitem[{Governo Brasileiro}(2024)]%
        {dadosabertos}
\bibfield{author}{\bibinfo{person}{{Governo Brasileiro}}.}
  \bibinfo{year}{2024}\natexlab{}.
\newblock \bibinfo{title}{{Portal Brasileiro de Dados Abertos}}.
\newblock \bibinfo{howpublished}{\url{https://dados.gov.br}}.
\newblock


\bibitem[Hegselmann et~al\mbox{.}(2023)]%
        {Hegselmann2022TabLLMFC}
\bibfield{author}{\bibinfo{person}{Stefan Hegselmann},
  \bibinfo{person}{Alejandro Buendia}, \bibinfo{person}{Hunter Lang},
  \bibinfo{person}{Monica Agrawal}, \bibinfo{person}{Xiaoyi Jiang}, {and}
  \bibinfo{person}{David Sontag}.} \bibinfo{year}{2023}\natexlab{}.
\newblock \showarticletitle{TabLLM: Few-shot Classification of Tabular Data
  with Large Language Models}. In \bibinfo{booktitle}{\emph{Proceedings of The
  International Conference on Artificial Intelligence and Statistics}},
  Vol.~\bibinfo{volume}{206}. \bibinfo{pages}{5549--5581}.
\newblock


\bibitem[Hendrycks and Dietterich(2019)]%
        {Hendrycks2019BenchmarkingNN}
\bibfield{author}{\bibinfo{person}{Dan Hendrycks} {and}
  \bibinfo{person}{Thomas~G. Dietterich}.} \bibinfo{year}{2019}\natexlab{}.
\newblock \showarticletitle{Benchmarking Neural Network Robustness to Common
  Corruptions and Perturbations}. In \bibinfo{booktitle}{\emph{International
  Conference on Learning Representations, {ICLR}}}.
  \bibinfo{publisher}{OpenReview.net}.
\newblock
\urldef\tempurl%
\url{https://openreview.net/forum?id=HJz6tiCqYm}
\showURL{%
\tempurl}


\bibitem[Hu et~al\mbox{.}(2019)]%
        {viznet}
\bibfield{author}{\bibinfo{person}{Kevin Hu}, \bibinfo{person}{Neil Gaikwad},
  \bibinfo{person}{Michiel Bakker}, \bibinfo{person}{Madelon Hulsebos},
  \bibinfo{person}{Emanuel Zgraggen}, \bibinfo{person}{C\'{e}sar Hidalgo},
  \bibinfo{person}{Tim Kraska}, \bibinfo{person}{Guoliang Li},
  \bibinfo{person}{Arvind Satyanarayan}, {and}
  \bibinfo{person}{{\c{C}}a{\u{g}}atay Demiralp}.}
  \bibinfo{year}{2019}\natexlab{}.
\newblock \showarticletitle{VizNet: {T}owards a large-scale visualization
  learning and benchmarking repository}. In
  \bibinfo{booktitle}{\emph{Proceedings of the Conference on Human Factors in
  Computing Systems (CHI)}}. \bibinfo{publisher}{ACM}, \bibinfo{pages}{1--12}.
\newblock


\bibitem[Hulsebos et~al\mbox{.}(2023)]%
        {Hulsebos2021GitTablesAL}
\bibfield{author}{\bibinfo{person}{Madelon Hulsebos},
  \bibinfo{person}{{\c{C}}agatay Demiralp}, {and} \bibinfo{person}{Paul
  Groth}.} \bibinfo{year}{2023}\natexlab{}.
\newblock \showarticletitle{Gittables: A large-scale corpus of relational
  tables}.
\newblock \bibinfo{journal}{\emph{Proceedings of the ACM on Management of
  Data}} \bibinfo{volume}{1}, \bibinfo{number}{1} (\bibinfo{year}{2023}),
  \bibinfo{pages}{1--17}.
\newblock


\bibitem[Hulsebos et~al\mbox{.}(2019)]%
        {sherlockpaper}
\bibfield{author}{\bibinfo{person}{Madelon Hulsebos}, \bibinfo{person}{Kevin
  Hu}, \bibinfo{person}{Michiel Bakker}, \bibinfo{person}{Emanuel Zgraggen},
  \bibinfo{person}{Arvind Satyanarayan}, \bibinfo{person}{Tim Kraska},
  \bibinfo{person}{\c{C}agatay Demiralp}, {and} \bibinfo{person}{C{\'e}sar
  Hidalgo}.} \bibinfo{year}{2019}\natexlab{}.
\newblock \showarticletitle{Sherlock: A Deep Learning Approach to Semantic Data
  Type Detection}. In \bibinfo{booktitle}{\emph{Proceedings of the ACM SIGKDD
  International Conference on Knowledge Discovery \& Data Mining}}.
  \bibinfo{publisher}{ACM}, \bibinfo{pages}{468--479}.
\newblock


\bibitem[Ilyas and Chu(2019)]%
        {chu2019data}
\bibfield{author}{\bibinfo{person}{Ihab~F. Ilyas} {and} \bibinfo{person}{Xu
  Chu}.} \bibinfo{year}{2019}\natexlab{}.
\newblock \bibinfo{booktitle}{\emph{Data Cleaning}}.
\newblock \bibinfo{publisher}{ACM}.
\newblock
\showISBNx{9781450371520}


\bibitem[Iyer et~al\mbox{.}(2022)]%
        {Iyer2022OPTIMLSL}
\bibfield{author}{\bibinfo{person}{Srinivasan Iyer},
  \bibinfo{person}{Xi~Victoria Lin}, \bibinfo{person}{Ramakanth Pasunuru},
  \bibinfo{person}{Todor Mihaylov}, \bibinfo{person}{Daniel Simig},
  \bibinfo{person}{Ping Yu}, \bibinfo{person}{Kurt Shuster},
  \bibinfo{person}{Tianlu Wang}, \bibinfo{person}{Qing Liu},
  \bibinfo{person}{Punit~Singh Koura}, \bibinfo{person}{Xian Li},
  \bibinfo{person}{Brian O'Horo}, \bibinfo{person}{Gabriel Pereyra},
  \bibinfo{person}{Jeff Wang}, \bibinfo{person}{Christopher Dewan},
  \bibinfo{person}{Asli Celikyilmaz}, \bibinfo{person}{Luke Zettlemoyer}, {and}
  \bibinfo{person}{Ves Stoyanov}.} \bibinfo{year}{2022}\natexlab{}.
\newblock \showarticletitle{{OPT-IML:} Scaling Language Model Instruction Meta
  Learning through the Lens of Generalization}.
\newblock \bibinfo{journal}{\emph{CoRR}}  \bibinfo{volume}{abs/2212.12017}
  (\bibinfo{year}{2022}).
\newblock
\urldef\tempurl%
\url{https://doi.org/10.48550/ARXIV.2212.12017}
\showDOI{\tempurl}
\showeprint[arXiv]{2212.12017}


\bibitem[Kandel et~al\mbox{.}(2011)]%
        {kandel2011wrangler}
\bibfield{author}{\bibinfo{person}{Sean Kandel}, \bibinfo{person}{Andreas
  Paepcke}, \bibinfo{person}{Joseph Hellerstein}, {and}
  \bibinfo{person}{Jeffrey Heer}.} \bibinfo{year}{2011}\natexlab{}.
\newblock \showarticletitle{Wrangler: Interactive visual specification of data
  transformation scripts}. In \bibinfo{booktitle}{\emph{Proceedings of the
  {SIGCHI} conference on human factors in computing systems}}.
  \bibinfo{publisher}{ACM}, \bibinfo{pages}{3363--3372}.
\newblock


\bibitem[Kayali et~al\mbox{.}(2023)]%
        {kayali2023chorus}
\bibfield{author}{\bibinfo{person}{Moe Kayali}, \bibinfo{person}{Anton Lykov},
  \bibinfo{person}{Ilias Fountalis}, \bibinfo{person}{Nikolaos Vasiloglou},
  \bibinfo{person}{Dan Olteanu}, {and} \bibinfo{person}{Dan Suciu}.}
  \bibinfo{year}{2023}\natexlab{}.
\newblock \showarticletitle{CHORUS: foundation models for unified data
  discovery and exploration}.
\newblock \bibinfo{journal}{\emph{arXiv preprint arXiv:2306.09610}}
  (\bibinfo{year}{2023}).
\newblock


\bibitem[Khatiwada et~al\mbox{.}(2023)]%
        {khatiwada2023santos}
\bibfield{author}{\bibinfo{person}{Aamod Khatiwada}, \bibinfo{person}{Grace
  Fan}, \bibinfo{person}{Roee Shraga}, \bibinfo{person}{Zixuan Chen},
  \bibinfo{person}{Wolfgang Gatterbauer}, \bibinfo{person}{Ren{\'e}e~J Miller},
  {and} \bibinfo{person}{Mirek Riedewald}.} \bibinfo{year}{2023}\natexlab{}.
\newblock \showarticletitle{SANTOS: Relationship-based Semantic Table Union
  Search}.
\newblock \bibinfo{journal}{\emph{Proceedings of the ACM on Management of
  Data}} \bibinfo{volume}{1}, \bibinfo{number}{1} (\bibinfo{year}{2023}),
  \bibinfo{pages}{1--25}.
\newblock


\bibitem[Korini and Bizer(2023)]%
        {korini2023column}
\bibfield{author}{\bibinfo{person}{Keti Korini} {and}
  \bibinfo{person}{Christian Bizer}.} \bibinfo{year}{2023}\natexlab{}.
\newblock \showarticletitle{Column type annotation using chatgpt}.
\newblock \bibinfo{journal}{\emph{arXiv preprint arXiv:2306.00745}}
  (\bibinfo{year}{2023}).
\newblock


\bibitem[Korini et~al\mbox{.}(2022)]%
        {sotab}
\bibfield{author}{\bibinfo{person}{Keti Korini}, \bibinfo{person}{Ralph
  Peeters}, {and} \bibinfo{person}{Christian Bizer}.}
  \bibinfo{year}{2022}\natexlab{}.
\newblock \showarticletitle{SOTAB: The WDC Schema. org table annotation
  benchmark}. In \bibinfo{booktitle}{\emph{CEUR Workshop Proceedings}},
  Vol.~\bibinfo{volume}{3320}. RWTH Aachen, \bibinfo{publisher}{Sun SITE
  Central Europe}, \bibinfo{pages}{14--19}.
\newblock


\bibitem[Miller et~al\mbox{.}(2021)]%
        {Miller2021AccuracyOT}
\bibfield{author}{\bibinfo{person}{John~P Miller}, \bibinfo{person}{Rohan
  Taori}, \bibinfo{person}{Aditi Raghunathan}, \bibinfo{person}{Shiori Sagawa},
  \bibinfo{person}{Pang~Wei Koh}, \bibinfo{person}{Vaishaal Shankar},
  \bibinfo{person}{Percy Liang}, \bibinfo{person}{Yair Carmon}, {and}
  \bibinfo{person}{Ludwig Schmidt}.} \bibinfo{year}{2021}\natexlab{}.
\newblock \showarticletitle{Accuracy on the line: on the strong correlation
  between out-of-distribution and in-distribution generalization}. In
  \bibinfo{booktitle}{\emph{International Conference on Machine Learning}}.
  PMLR, \bibinfo{pages}{7721--7735}.
\newblock


\bibitem[Muennighoff et~al\mbox{.}(2023)]%
        {Muennighoff2022CrosslingualGT}
\bibfield{author}{\bibinfo{person}{Niklas Muennighoff}, \bibinfo{person}{Thomas
  Wang}, \bibinfo{person}{Lintang Sutawika}, \bibinfo{person}{Adam Roberts},
  \bibinfo{person}{Stella Biderman}, \bibinfo{person}{Teven Le~Scao},
  \bibinfo{person}{M~Saiful Bari}, \bibinfo{person}{Sheng Shen},
  \bibinfo{person}{Zheng~Xin Yong}, \bibinfo{person}{Hailey Schoelkopf},
  \bibinfo{person}{Xiangru Tang}, \bibinfo{person}{Dragomir Radev},
  \bibinfo{person}{Alham~Fikri Aji}, \bibinfo{person}{Khalid Almubarak},
  \bibinfo{person}{Samuel Albanie}, \bibinfo{person}{Zaid Alyafeai},
  \bibinfo{person}{Albert Webson}, \bibinfo{person}{Edward Raff}, {and}
  \bibinfo{person}{Colin Raffel}.} \bibinfo{year}{2023}\natexlab{}.
\newblock \showarticletitle{Crosslingual Generalization through Multitask
  Finetuning}. In \bibinfo{booktitle}{\emph{Proceedings of the Annual Meeting
  of the Association for Computational Linguistics (Volume 1: Long Papers)}}.
  \bibinfo{publisher}{Association for Computational Linguistics},
  \bibinfo{pages}{15991--16111}.
\newblock


\bibitem[Narayan et~al\mbox{.}(2022)]%
        {Narayan2022CanFM}
\bibfield{author}{\bibinfo{person}{Avanika Narayan}, \bibinfo{person}{Ines
  Chami}, \bibinfo{person}{Laurel~J. Orr}, {and} \bibinfo{person}{Christopher
  R\'e}.} \bibinfo{year}{2022}\natexlab{}.
\newblock \showarticletitle{Can Foundation Models Wrangle Your Data?}
\newblock \bibinfo{journal}{\emph{Proc. VLDB Endow.}}  \bibinfo{volume}{16}
  (\bibinfo{year}{2022}), \bibinfo{pages}{738--746}.
\newblock


\bibitem[{NYC Office of Technology and Innovation (OTI)}(2024)]%
        {nyc_opendata}
\bibfield{author}{\bibinfo{person}{{NYC Office of Technology and Innovation
  (OTI)}}.} \bibinfo{year}{2024}\natexlab{}.
\newblock \bibinfo{title}{{NYC Open Data}}.
\newblock
\newblock


\bibitem[Opitz and Frank(2022)]%
        {Opitz2022SBERTSM}
\bibfield{author}{\bibinfo{person}{Juri Opitz} {and} \bibinfo{person}{Anette
  Frank}.} \bibinfo{year}{2022}\natexlab{}.
\newblock \showarticletitle{SBERT studies meaning representations: Decomposing
  sentence embeddings into explainable semantic features}. In
  \bibinfo{booktitle}{\emph{Proceedings of the Conference of the Asia-Pacific
  Chapter of the Association for Computational Linguistics and the 12th
  International Joint Conference on Natural Language Processing}}.
  \bibinfo{publisher}{Association for Computational Linguistics},
  \bibinfo{pages}{625--638}.
\newblock


\bibitem[Ota et~al\mbox{.}(2020)]%
        {D4paper}
\bibfield{author}{\bibinfo{person}{Masayo Ota}, \bibinfo{person}{Heiko
  M\"{u}ller}, \bibinfo{person}{Juliana Freire}, {and} \bibinfo{person}{Divesh
  Srivastava}.} \bibinfo{year}{2020}\natexlab{}.
\newblock \showarticletitle{Data-Driven Domain Discovery for Structured
  Datasets}.
\newblock \bibinfo{journal}{\emph{Proc. VLDB Endow.}} \bibinfo{volume}{13},
  \bibinfo{number}{7} (\bibinfo{date}{mar} \bibinfo{year}{2020}),
  \bibinfo{pages}{953–967}.
\newblock
\showISSN{2150-8097}


\bibitem[Ouyang et~al\mbox{.}(2022)]%
        {Ouyang2022TrainingLM}
\bibfield{author}{\bibinfo{person}{Long Ouyang}, \bibinfo{person}{Jeffrey Wu},
  \bibinfo{person}{Xu Jiang}, \bibinfo{person}{Diogo Almeida},
  \bibinfo{person}{Carroll Wainwright}, \bibinfo{person}{Pamela Mishkin},
  \bibinfo{person}{Chong Zhang}, \bibinfo{person}{Sandhini Agarwal},
  \bibinfo{person}{Katarina Slama}, \bibinfo{person}{Alex Ray},
  \bibinfo{person}{John Schulman}, \bibinfo{person}{Jacob Hilton},
  \bibinfo{person}{Fraser Kelton}, \bibinfo{person}{Luke Miller},
  \bibinfo{person}{Maddie Simens}, \bibinfo{person}{Amanda Askell},
  \bibinfo{person}{Peter Welinder}, \bibinfo{person}{Paul~F Christiano},
  \bibinfo{person}{Jan Leike}, {and} \bibinfo{person}{Ryan Lowe}.}
  \bibinfo{year}{2022}\natexlab{}.
\newblock \showarticletitle{Training language models to follow instructions
  with human feedback}. In \bibinfo{booktitle}{\emph{Advances in Neural
  Information Processing Systems}}, Vol.~\bibinfo{volume}{35}.
  \bibinfo{pages}{27730--27744}.
\newblock


\bibitem[Power et~al\mbox{.}(2022)]%
        {Power2022GrokkingGB}
\bibfield{author}{\bibinfo{person}{Alethea Power}, \bibinfo{person}{Yuri
  Burda}, \bibinfo{person}{Harri Edwards}, \bibinfo{person}{Igor Babuschkin},
  {and} \bibinfo{person}{Vedant Misra}.} \bibinfo{year}{2022}\natexlab{}.
\newblock \showarticletitle{Grokking: Generalization beyond overfitting on
  small algorithmic datasets}.
\newblock \bibinfo{journal}{\emph{arXiv preprint arXiv:2201.02177}}
  (\bibinfo{year}{2022}).
\newblock


\bibitem[Pradeep et~al\mbox{.}(2023)]%
        {pradeep2023rankvicuna}
\bibfield{author}{\bibinfo{person}{Ronak Pradeep}, \bibinfo{person}{Sahel
  Sharifymoghaddam}, {and} \bibinfo{person}{Jimmy Lin}.}
  \bibinfo{year}{2023}\natexlab{}.
\newblock \showarticletitle{Rankvicuna: Zero-shot listwise document reranking
  with open-source large language models}.
\newblock \bibinfo{journal}{\emph{arXiv preprint arXiv:2309.15088}}
  (\bibinfo{year}{2023}).
\newblock


\bibitem[Quinonero-Candela et~al\mbox{.}(2008)]%
        {quinonero2008dataset}
\bibfield{author}{\bibinfo{person}{Joaquin Quinonero-Candela},
  \bibinfo{person}{Masashi Sugiyama}, \bibinfo{person}{Anton Schwaighofer},
  {and} \bibinfo{person}{Neil~D Lawrence}.} \bibinfo{year}{2008}\natexlab{}.
\newblock \bibinfo{booktitle}{\emph{Dataset shift in machine learning}}.
\newblock \bibinfo{publisher}{Mit Press}.
\newblock


\bibitem[Radford et~al\mbox{.}(2021)]%
        {Radford2021LearningTV}
\bibfield{author}{\bibinfo{person}{Alec Radford}, \bibinfo{person}{Jong~Wook
  Kim}, \bibinfo{person}{Chris Hallacy}, \bibinfo{person}{Aditya Ramesh},
  \bibinfo{person}{Gabriel Goh}, \bibinfo{person}{Sandhini Agarwal},
  \bibinfo{person}{Girish Sastry}, \bibinfo{person}{Amanda Askell},
  \bibinfo{person}{Pamela Mishkin}, \bibinfo{person}{Jack Clark},
  \bibinfo{person}{Gretchen Krueger}, {and} \bibinfo{person}{Ilya Sutskever}.}
  \bibinfo{year}{2021}\natexlab{}.
\newblock \showarticletitle{Learning Transferable Visual Models From Natural
  Language Supervision}. In \bibinfo{booktitle}{\emph{ICML}}.
  \bibinfo{pages}{1090--1094}.
\newblock


\bibitem[Raman and Hellerstein(2001)]%
        {raman2001potter}
\bibfield{author}{\bibinfo{person}{Vijayshankar Raman} {and}
  \bibinfo{person}{Joseph~M Hellerstein}.} \bibinfo{year}{2001}\natexlab{}.
\newblock \showarticletitle{Potter's wheel: An interactive data cleaning
  system}. In \bibinfo{booktitle}{\emph{VLDB}}, Vol.~\bibinfo{volume}{1}.
  \bibinfo{pages}{381--390}.
\newblock


\bibitem[Recht et~al\mbox{.}(2019)]%
        {Recht2019DoIC}
\bibfield{author}{\bibinfo{person}{Benjamin Recht}, \bibinfo{person}{Rebecca
  Roelofs}, \bibinfo{person}{Ludwig Schmidt}, {and} \bibinfo{person}{Vaishaal
  Shankar}.} \bibinfo{year}{2019}\natexlab{}.
\newblock \showarticletitle{Do imagenet classifiers generalize to imagenet?}.
  In \bibinfo{booktitle}{\emph{International conference on machine learning}}.
  PMLR, \bibinfo{publisher}{ICML}, \bibinfo{pages}{5389--5400}.
\newblock


\bibitem[Rogers et~al\mbox{.}(2023)]%
        {rogers-etal-2023-closed}
\bibfield{author}{\bibinfo{person}{Anna Rogers}, \bibinfo{person}{Niranjan
  Balasubramanian}, \bibinfo{person}{Leon Derczynski}, \bibinfo{person}{Jesse
  Dodge}, \bibinfo{person}{Alexander Koller}, \bibinfo{person}{Sasha Luccioni},
  \bibinfo{person}{Maarten Sap}, \bibinfo{person}{Roy Schwartz},
  \bibinfo{person}{Noah~A Smith}, {and} \bibinfo{person}{Emma Strubell}.}
  \bibinfo{year}{2023}\natexlab{}.
\newblock \showarticletitle{Closed ai models make bad baselines}.
\newblock \bibinfo{journal}{\emph{Hacking Semantics}}  \bibinfo{volume}{3}
  (\bibinfo{year}{2023}).
\newblock


\bibitem[Sclar et~al\mbox{.}(2023)]%
        {sclar_quantifying_2023}
\bibfield{author}{\bibinfo{person}{Melanie Sclar}, \bibinfo{person}{Yejin
  Choi}, \bibinfo{person}{Yulia Tsvetkov}, {and} \bibinfo{person}{Alane Suhr}.}
  \bibinfo{year}{2023}\natexlab{}.
\newblock \showarticletitle{Quantifying Language Models' Sensitivity to
  Spurious Features in Prompt Design or: How I learned to start worrying about
  prompt formatting}.
\newblock \bibinfo{journal}{\emph{arXiv preprint arXiv:2310.11324}}
  (\bibinfo{year}{2023}).
\newblock


\bibitem[Sennrich et~al\mbox{.}(2016)]%
        {Sennrich2015NeuralMT}
\bibfield{author}{\bibinfo{person}{Rico Sennrich}, \bibinfo{person}{Barry
  Haddow}, {and} \bibinfo{person}{Alexandra Birch}.}
  \bibinfo{year}{2016}\natexlab{}.
\newblock \showarticletitle{Neural Machine Translation of Rare Words with
  Subword Units}. In \bibinfo{booktitle}{\emph{Proceedings of the Annual
  Meeting of the Association for Computational Linguistics (Volume 1: Long
  Papers)}}. \bibinfo{publisher}{Association for Computational Linguistics},
  \bibinfo{pages}{1715--1725}.
\newblock


\bibitem[Suhara et~al\mbox{.}(2022)]%
        {doduopaper}
\bibfield{author}{\bibinfo{person}{Yoshihiko Suhara}, \bibinfo{person}{Jinfeng
  Li}, \bibinfo{person}{Yuliang Li}, \bibinfo{person}{Dan Zhang},
  \bibinfo{person}{\c{C}a\u{g}atay Demiralp}, \bibinfo{person}{Chen Chen},
  {and} \bibinfo{person}{Wang-Chiew Tan}.} \bibinfo{year}{2022}\natexlab{}.
\newblock \showarticletitle{Annotating Columns with Pre-Trained Language
  Models}. In \bibinfo{booktitle}{\emph{Proceedings of the International
  Conference on Management of Data (SIGMOD)}}. \bibinfo{publisher}{ACM},
  \bibinfo{pages}{1493–1503}.
\newblock


\bibitem[Sutton(2019)]%
        {sutton2019bitter}
\bibfield{author}{\bibinfo{person}{Richard~S. Sutton}.}
  \bibinfo{year}{2019}\natexlab{}.
\newblock \bibinfo{title}{The Bitter Lesson}.
\newblock
\newblock


\bibitem[Taori et~al\mbox{.}(2023)]%
        {alpaca}
\bibfield{author}{\bibinfo{person}{Rohan Taori}, \bibinfo{person}{Ishaan
  Gulrajani}, \bibinfo{person}{Tianyi Zhang}, \bibinfo{person}{Yann Dubois},
  \bibinfo{person}{Xuechen Li}, \bibinfo{person}{Carlos Guestrin},
  \bibinfo{person}{Percy Liang}, {and} \bibinfo{person}{Tatsunori~B.
  Hashimoto}.} \bibinfo{year}{2023}\natexlab{}.
\newblock \bibinfo{title}{Stanford Alpaca: An Instruction-following LLaMA
  model}.
\newblock
  \bibinfo{howpublished}{\url{https://github.com/tatsu-lab/stanford_alpaca}}.
\newblock


\bibitem[Tay et~al\mbox{.}(2023)]%
        {Tay2022UL2UL}
\bibfield{author}{\bibinfo{person}{Yi Tay}, \bibinfo{person}{Mostafa Dehghani},
  \bibinfo{person}{Vinh~Q. Tran}, \bibinfo{person}{Xavier Garcia},
  \bibinfo{person}{Jason Wei}, \bibinfo{person}{Xuezhi Wang},
  \bibinfo{person}{Hyung~Won Chung}, \bibinfo{person}{Dara Bahri},
  \bibinfo{person}{Tal Schuster}, \bibinfo{person}{Huaixiu~Steven Zheng},
  \bibinfo{person}{Denny Zhou}, \bibinfo{person}{Neil Houlsby}, {and}
  \bibinfo{person}{Donald Metzler}.} \bibinfo{year}{2023}\natexlab{}.
\newblock \showarticletitle{{UL2:} Unifying Language Learning Paradigms}. In
  \bibinfo{booktitle}{\emph{The Eleventh International Conference on Learning
  Representations, {ICLR}}}. \bibinfo{publisher}{OpenReview.net}.
\newblock
\urldef\tempurl%
\url{https://openreview.net/pdf?id=6ruVLB727MC}
\showURL{%
\tempurl}


\bibitem[Touvron et~al\mbox{.}(2023a)]%
        {Touvron2023LLaMAOA}
\bibfield{author}{\bibinfo{person}{Hugo Touvron}, \bibinfo{person}{Thibaut
  Lavril}, \bibinfo{person}{Gautier Izacard}, \bibinfo{person}{Xavier
  Martinet}, \bibinfo{person}{Marie-Anne Lachaux},
  \bibinfo{person}{Timoth{\'e}e Lacroix}, \bibinfo{person}{Baptiste
  Rozi{\`e}re}, \bibinfo{person}{Naman Goyal}, \bibinfo{person}{Eric Hambro},
  \bibinfo{person}{Faisal Azhar}, {et~al\mbox{.}}}
  \bibinfo{year}{2023}\natexlab{a}.
\newblock \showarticletitle{Llama: Open and efficient foundation language
  models}.
\newblock \bibinfo{journal}{\emph{arXiv preprint arXiv:2302.13971}}
  (\bibinfo{year}{2023}).
\newblock


\bibitem[Touvron et~al\mbox{.}(2023b)]%
        {touvron2023llama}
\bibfield{author}{\bibinfo{person}{Hugo Touvron}, \bibinfo{person}{Louis
  Martin}, \bibinfo{person}{Kevin Stone}, \bibinfo{person}{Peter Albert},
  \bibinfo{person}{Amjad Almahairi}, \bibinfo{person}{Yasmine Babaei},
  \bibinfo{person}{Nikolay Bashlykov}, \bibinfo{person}{Soumya Batra},
  \bibinfo{person}{Prajjwal Bhargava}, \bibinfo{person}{Shruti Bhosale},
  {et~al\mbox{.}}} \bibinfo{year}{2023}\natexlab{b}.
\newblock \showarticletitle{Llama 2: Open foundation and fine-tuned chat
  models}.
\newblock \bibinfo{journal}{\emph{arXiv preprint arXiv:2307.09288}}
  (\bibinfo{year}{2023}).
\newblock


\bibitem[Tu et~al\mbox{.}(2023)]%
        {tu_unicorn_2023}
\bibfield{author}{\bibinfo{person}{Jianhong Tu}, \bibinfo{person}{Ju Fan},
  \bibinfo{person}{Nan Tang}, \bibinfo{person}{Peng Wang},
  \bibinfo{person}{Guoliang Li}, \bibinfo{person}{Xiaoyong Du},
  \bibinfo{person}{Xiaofeng Jia}, {and} \bibinfo{person}{Song Gao}.}
  \bibinfo{year}{2023}\natexlab{}.
\newblock \showarticletitle{Unicorn: {A} {Unified} {Multi}-tasking {Model} for
  {Supporting} {Matching} {Tasks} in {Data} {Integration}}.
\newblock \bibinfo{journal}{\emph{Proceedings of the ACM on Management of
  Data}} \bibinfo{volume}{1}, \bibinfo{number}{1} (\bibinfo{year}{2023}),
  \bibinfo{pages}{1--26}.
\newblock
\urldef\tempurl%
\url{https://doi.org/10.1145/3588938}
\showDOI{\tempurl}


\bibitem[Vaswani et~al\mbox{.}(2017)]%
        {transformer}
\bibfield{author}{\bibinfo{person}{Ashish Vaswani}, \bibinfo{person}{Noam
  Shazeer}, \bibinfo{person}{Niki Parmar}, \bibinfo{person}{Jakob Uszkoreit},
  \bibinfo{person}{Llion Jones}, \bibinfo{person}{Aidan~N. Gomez},
  \bibinfo{person}{\L{}ukasz Kaiser}, {and} \bibinfo{person}{Illia
  Polosukhin}.} \bibinfo{year}{2017}\natexlab{}.
\newblock \showarticletitle{Attention is All You Need}. In
  \bibinfo{booktitle}{\emph{Proceedings of the International Conference on
  Neural Information Processing Systems (NEURIPS)}}.
  \bibinfo{pages}{5998--6008}.
\newblock


\bibitem[Wolf et~al\mbox{.}(2019)]%
        {huggingface_transformers}
\bibfield{author}{\bibinfo{person}{Thomas Wolf}, \bibinfo{person}{Lysandre
  Debut}, \bibinfo{person}{Victor Sanh}, \bibinfo{person}{Julien Chaumond},
  \bibinfo{person}{Clement Delangue}, \bibinfo{person}{Anthony Moi},
  \bibinfo{person}{Pierric Cistac}, \bibinfo{person}{Tim Rault},
  \bibinfo{person}{R{\'e}mi Louf}, \bibinfo{person}{Morgan Funtowicz},
  {et~al\mbox{.}}} \bibinfo{year}{2019}\natexlab{}.
\newblock \showarticletitle{Huggingface's transformers: State-of-the-art
  natural language processing}.
\newblock \bibinfo{journal}{\emph{arXiv preprint arXiv:1910.03771}}
  (\bibinfo{year}{2019}).
\newblock


\bibitem[Zhang et~al\mbox{.}(2020)]%
        {zhang2020sato}
\bibfield{author}{\bibinfo{person}{Dan Zhang}, \bibinfo{person}{Yoshihiko
  Suhara}, \bibinfo{person}{Jinfeng Li}, \bibinfo{person}{Madelon Hulsebos},
  \bibinfo{person}{{\c{C}}a{\u{g}}atay Demiralp}, {and}
  \bibinfo{person}{Wang-Chiew Tan}.} \bibinfo{year}{2020}\natexlab{}.
\newblock \showarticletitle{Sato: Contextual Semantic Type Detection in
  Tables}.
\newblock \bibinfo{journal}{\emph{Proc. VLDB Endow.}} \bibinfo{volume}{13},
  \bibinfo{number}{12} (\bibinfo{year}{2020}), \bibinfo{pages}{1835–1848}.
\newblock


\end{thebibliography}

\clearpage

\begin{appendix}
\section{How often do LLMs generate invalid labels?}
\label{app:llm-invalid}

\begin{table}[h]
    \centering
    \resizebox{\columnwidth}{!}{%
\begin{tabular}{lrrrrrrrr}
\textbf{Dataset Name} & \multicolumn{1}{l}{\textbf{\# Cols}} & \multicolumn{1}{l}{\textbf{RS1}} & \multicolumn{1}{l}{\textbf{RS2}} & \multicolumn{1}{l}{\textbf{RS3}} & \multicolumn{1}{l}{\textbf{RS4}} & \multicolumn{1}{l}{\textbf{RS5}} & \multicolumn{1}{l}{\textbf{RS Avg. Pct.}} & \multicolumn{1}{l}{\textbf{ZS Avg. Acc.}} \\
D4-20 & 2000 & 0 & 105 & 107 & 109 & 235 & 5.6 & 81.2 \\
PubChem-20 & 2000 & 1 & 30 & 109 & 201 & 531 & 8.7 & 65.3 \\
SOTAB-27 & 15040 & 85 & 256 & 729 & 2133 & 3960 & 9.5 & 57.4 \\
Amstr-56 & 2000 & 172 & 213 & 328 & 808 & 1429 & 29.5 & 19.7
\end{tabular}%
}
    \caption{\sl\textbf{Effects of label remapping. } \sl Label remapping is common on most zero-shot datasets, and inversely correlates with model performance. RS stands for random sample (from the space of all ArcheType-T5 runs on that dataset), ZS for zero shot, \# Cols for number of columns, Avg. Pct. for average percentage remapped (out of the total columns in the dataset).}
    \label{tab:label-remapping-freq}
\end{table}

\begin{table*}[!tb]
\resizebox{\textwidth}{!}{
\begin{tabular}{@{}lrrlr@{}}
\textbf{Class (A)} & \multicolumn{1}{l}{\textbf{T5 Acc. (A)}} & \multicolumn{1}{l}{\textbf{T5 Acc. (A, S)}} & \textbf{Class (B)} & \multicolumn{1}{l}{\textbf{T5 Acc. (B)}} \\
abstract for patent & 0.97 & \cellcolor[HTML]{F9CB9C}0.53 & abstract for patent & 0.99 \\
biological formula & 0 & 0 & \cellcolor[HTML]{FFFF00}iupac & 0 \\
book isbn & 1 & 1 & book isbn & 1 \\
book title & 0.06 & \cellcolor[HTML]{D9EAD3}0.26 & book title & 0.06 \\
cell alternative label & 0.75 & \cellcolor[HTML]{D9EAD3}0.88 & \cellcolor[HTML]{FFFF00}cell label & 0.76 \\
chemical & 0.64 & \cellcolor[HTML]{D9EAD3}0.86 & \cellcolor[HTML]{FFFF00}concept preferred label & \cellcolor[HTML]{F8CBAD}0.06 \\
concept broader term & 1 & 1 & concept broader term & 1 \\
disease alternative label & 1 & 1 & disease label & 1 \\
inchi (international chemical identifier) & 1 & 1 & inchi (international chemical identifier) & 1 \\
journal issn & 1 & 1 & journal issn & 1 \\
journal title & 1 & 1 & journal title & 1 \\
md5 hash & 1 & 1 & md5 hash & 1 \\
molecular formula & 0.99 & \cellcolor[HTML]{F9CB9C}0.24 & molecular formula & 1 \\
organization & 0.96 & 0.96 & organization & 0.98 \\
patent title & 0.96 & 1 & patent title & \cellcolor[HTML]{F8CBAD}0.68 \\
person's first name and middle initials & 0.05 & 0 & \cellcolor[HTML]{FFFF00}author first name & 0 \\
person's full name & 1 & 1 & \cellcolor[HTML]{FFFF00}author full name & 0.99 \\
person's last name & 0.78 & \cellcolor[HTML]{F8CBAD}0.59 & \cellcolor[HTML]{FFFF00}author family name & 0.79 \\
smiles (simplified molecular input line entry system) & 0.73 & \cellcolor[HTML]{F9CB9C}0.61 & smiles (simplified molecular input line entry system) & \cellcolor[HTML]{F8CBAD}0.49 \\
taxonomy label & 1 & 1 & taxonomy label & 1
\end{tabular}}
\caption{\sl\textbf{Choice of class labels has unpredictable effects across classes. } \sl Changing the label associated with a class will have unpredictable effects on performance; namely, performance of relabeled class may not change, while performances of classes with the \textit{same} labels \textit{does} change. Shuffling the order of classnames in the prompt (A, S) also affects accuracy. Results shown are zero-shot Micro-F1 scores on the Pubchem-20 dataset. For convenience, the changed classnames are highlighted in yellow, and significant (> 3\%) changes in per-class accuracy are higlighted in orange or green. Table best viewed in color.}
\label{tab:cn-ablation}
\end{table*}

As discussed in our main paper, generative models such as LLMs sometimes produce labels which are not in the target set. But how often does this occur? We find that the frequency of this event varies widely, depending on the model and the dataset.

By way of illustrating the high degree of variance, in \Tabref{tab:label-remapping-freq}, we report the number of samples remapped per dataset on five randomly selected experiments from our ablation studies; these random samples are fixed only with respect to the choice of pretraining dataset and the fact that they are all zero-shot. Therefore, they vary across the space of all architectures, prompts, sample sizes and remapping strategies which appear elsewhere in the paper. Modifying these conditions, we observe a high degree of variance between runs.

\paragraph{How do remapped labels affect model performance?} When we compare average model performance on a zero-shot benchmark to average number of remapped labels in \Tabref{tab:label-remapping-freq}, we see that they are inversely correlated -- the more labels remapped, the less accurate the model becomes, on average. 

This provides some useful intuition for why label remapping can have such a large effect on the overall performance of a method; when the LLM-CTA model encounters a challenging sample, it becomes more likely that (1) the LLM will generate an out-of-distribution answer, (2), the LLM will generate an incorrect in-distribution answer. Any label remapping method will, by definition, eliminate all entries in category (1). As these entries would be incorrect without remapping, label remapping can only improve model performance. Furthermore, the better the label remapping technique, the more it will improve model performance.

\section{Example of a rule-based remapping change}
\label{app:ex-rbrm}

As a reference for future researchers, we provide an example of a specific rule which we found led to performance gains. For a complete list, we refer the reader to \cite{penfever2023archetype}.

\begin{example*}
    We observe that the SOTAB-91 label set contains a label, $\texttt{URL}$, and that the columns labeled as such have elements that are valid URL strings, such as the following:

   \begin{verbatim}       "http://empirebar.com.au/8.6.19/file.html?
       is_for_sharing=true"
   \end{verbatim}

    \noindent The SOTAB-91 dataset contains other labels (\texttt{attendenum}, \\ \texttt{availabilityofitem}, \texttt{offeritemcondition}, \texttt{statustype})\\ with type Schema.org enumerationtype and whose elements are a small number of specific Schema.org URLs.

   \begin{verbatim}        "http://schema.org/OfflineEventAttendanceMode"
   \end{verbatim}

    \noindent The columns with the latter labels are degenerate, because $\texttt{URL}$ is an equally valid and semantically more plausible label than $\texttt{attendenum}$.
    We therefore apply a simple lookup mapping from the enumerationtype label to the corresponding set of Schema.org URLs.
    This rule leads to the following per-class accuracy changes on an ArcheType-LLAMA-7B model:
    \begin{verbatim}
        attendenum: 5% -> 100%
        availabilityofitem: 56% -> 73%
        formatofbook: 80% -> 82%
        offeritemcondition: 42% -> 81%
        statustype: 32% -> 89%
        url: 83% -> 81%
    \end{verbatim}
\end{example*}

\begin{table*}
    \centering
    \begin{minipage}[t]{0.48\textwidth}
    \resizebox{.97\columnwidth}{!}{
    \begin{tabular}{@{}lrrrrl@{}}
& \multicolumn{1}{l}{} & \multicolumn{1}{l}{} & \multicolumn{1}{l}{} &  \\
& \multicolumn{1}{l}{} & \multicolumn{1}{l}{} & \multicolumn{1}{l}{} &  \\
& \multicolumn{1}{l}{} & \multicolumn{1}{l}{} & \multicolumn{1}{l}{} &  \\
& \multicolumn{1}{l}{} & \multicolumn{1}{l}{} & \multicolumn{1}{l}{} &  \\
& \multicolumn{1}{l}{} & \multicolumn{1}{l}{} & \multicolumn{1}{l}{} &  \\
\textbf{Class} & \multicolumn{1}{l}{\textbf{freq}} & \multicolumn{1}{l}{\textbf{T5}} & \multicolumn{1}{l}{\textbf{UL2}} & \multicolumn{1}{l}{\textbf{GPT}} & \textbf{Conf. Cls.} \\
age & 27 & 0.74 & 1 & 0.44 & category \\
boolean & 269 & 0.98 & 0.95 & 0.95 &  \\
category & 1437 & 0.15 & 0.14 & 0.6 & person, product, text \\
company & 726 & 0.52 & 0.51 & 0.21 & organization, streetaddress \\
coordinates & 191 & 0 & 0.99 & 0.77 & number \\
country & 413 & 0.78 & 0.74 & 0.43 & category, streetaddress \\
creativework & 1147 & 0.77 & 0.44 & 0.82 & event, product \\
currency & 280 & 0.97 & 0.97 & 0.91 &  \\
date & 867 & 0.63 & 0.55 & 0.83 & time \\
email & 140 & 0.66 & 0.35 & 0.97 & url \\
event & 422 & 0.73 & 0.9 & 0.51 & creativework \\
gender & 183 & 0.99 & 0.49 & 0.67 & person \\
jobposting & 13 & 0.92 & 0.62 & 0.77 & creativework, organization \\
jobrequirements & 167 & 0.01 & 0 & 0.01 & jobposting \\
language & 252 & 1 & 0.98 & 0.77 & text \\
number & 1417 & 0.65 & 0.5 & 0.61 & product, coordinates \\
organization & 758 & 0.33 & 0.36 & 0.37 & company, streetaddress \\
person & 606 & 0.71 & 0.79 & 0.67 & organization \\
price & 574 & 0.39 & 0.58 & 0.55 & currency, number \\
product & 622 & 0.63 & 0.63 & 0.63 & company \\
sportsteam & 51 & 0.82 & 0.86 & 0.69 & organization, person \\
streetaddress & 704 & 0.53 & 0.76 & 0.89 & country \\
telephone & 474 & 0.89 & 0.92 & 0.94 &  \\
text & 1289 & 0.44 & 0.29 & 0.36 & product, event \\
time & 807 & 0.75 & 0.71 & 0.84 &  \\
url & 460 & 0.85 & 0.91 & 0.85 &  \\
weight & 547 & 0.65 & 0.66 & 0.62 & coordinates \\
zipcode & 197 & 0.58 & 0.14 & 0.61 & streetaddress, number
\end{tabular}}
    \caption{\sl\textbf{Per class accuracy scores on the SOTAB-27 benchmark. } SOTAB-27 contains a mix of high-level abstract types (category, boolean) and low-level semantic and numeric types (country, sportsteam). We find that LLMs tend to favor the latter over the former, and this phenomenon is more pronounced on open-source models. LLM classification on PubChem exhibits a high degree of bias in favor of certain semantically distinct classes (job posting) over others (job requirements) -- the capacity to distinguish between closely related semantic types zero-shot is limited in current-generation LLMs. Conf. Cls. is a list of classes which were commonly confused with that target class.}
    \label{tab:sotab-class}
    \end{minipage}
    \hfill
    \begin{minipage}[t]{0.48\textwidth}
    \resizebox{.95\columnwidth}{!}{
    \begin{tabular}{@{}lrrrl@{}}
 % & \multicolumn{1}{l}{} & \multicolumn{1}{l}{} & \multicolumn{1}{l}{} &  \\
 % & \multicolumn{1}{l}{} & \multicolumn{1}{l}{} & \multicolumn{1}{l}{} &  \\
 % & \multicolumn{1}{l}{} & \multicolumn{1}{l}{} & \multicolumn{1}{l}{} &  \\
 % & \multicolumn{1}{l}{} & \multicolumn{1}{l}{} & \multicolumn{1}{l}{} &  \\
 % & \multicolumn{1}{l}{} & \multicolumn{1}{l}{} & \multicolumn{1}{l}{} &  \\
 % & \multicolumn{1}{l}{} & \multicolumn{1}{l}{} & \multicolumn{1}{l}{} &  \\
 % & \multicolumn{1}{l}{} & \multicolumn{1}{l}{} & \multicolumn{1}{l}{} &  \\
 % & \multicolumn{1}{l}{} & \multicolumn{1}{l}{} & \multicolumn{1}{l}{} &  \\
 % & \multicolumn{1}{l}{} & \multicolumn{1}{l}{} & \multicolumn{1}{l}{} &  \\
\textbf{Class} & \multicolumn{1}{l}{\textbf{T5}} & \multicolumn{1}{l}{\textbf{UL2}} & \multicolumn{1}{l}{\textbf{GPT}} & \textbf{Conf. Cls.} \\
abbreviation of agency & 0.87 & 0.71 & 0.02 & nyc agency name \\
borough & 1 & 1 & 1 &  \\
color & 1 & 1 & 1 &  \\
elevator or staircase & 1 & 1 & 0.48 & nyc agency name \\
ethnicity & 0.99 & 0.99 & 0.99 &  \\
month & 0.99 & 0.99 & 0.99 &  \\
nyc agency name & 1 & 1 & 1 &  \\
other-states & 1 & 1 & 1 &  \\
permit-types & 0.79 & 0.79 & 0.79 & other states \\
plate-type & 1 & 1 & 1 &  \\
region in bronx & 0.46 & 0.23 & 0.9 & other regions \\
region in brooklyn & 0.26 & 0.88 & 0.95 & other regions \\
region in manhattan & 0.99 & 0.68 & 0.99 & other regions \\
region in queens & 0.94 & 0.35 & 0.66 & other regions \\
region in staten island & 0 & 0 & 0.87 & other regions \\
school name & 1 & 1 & 1 &  \\
school-dbn & 1 & 1 & 1 &  \\
school-grades & 1 & 1 & 1 &  \\
school-number & 1 & 1 & 1 &  \\
us-state & 1 & 1 & 1 & 
\end{tabular}}
    \caption{\sl\textbf{Per class accuracy scores on the D4-20 benchmark. } We observe a high degree of consistency across LLMs in their handling of most classes. LLM classification on D4Tables exhibits a high degree of bias in favor of certain semantically distinct classes (nyc agency name) over others (abbreviation of agency) -- the capacity to distinguish between closely related semantic types zero-shot is limited in current-generation LLMs. GPT's larger pretraining corpus results in significantly better performance on region-specific classifications. Conf. Cls. is a list of classes which were commonly confused with that target class.}
    \label{tab:d4-class}
    \end{minipage}
\end{table*}

\section{Ablation on classname semantics and position}
\label{app:cn-sem-pos}

\myparagraphemph{Semantic changes to label names can have unpredictable effects on performance.} In \Tabref{tab:cn-ablation}, we introduce two label sets, termed A and B, for the PubChem-20 dataset. Label set B contains 6 semantically changed labels, compared to label set A. When we run the experiment with Label set B and compare the results to label set A, we find substantial accuracy changes to 3 classes; however, only 1 of these classes was among the 6 labels we changed.

From this experiment, we conclude the following: (1), contemporary LLMs are sensitive to changes in the label space. (2), this sensitivity is, at times, the functional equivalent of label noise, in that it produces seemingly random and unpredictable changes in test set accuracy. (3), the changes in performance are not confined to the class names modified, but distributed across the entire class space.

\myparagraphemph{Changes to label ordering can have unpredictable effects on performance.} In \Tabref{tab:cn-ablation}, for label set A, we also experiment with randomly shuffling the order in which classnames are presented to the model -- we call this experiment (A, S). For all other experiments reported in the paper, we sort classnames in ascending alphabetical order during serialization. We observe substantial changes in accuracy to 7 of 20 classes as a result of this transformation.

From this experiment, we conclude the following: (1), contemporary LLMs are sensitive to changes in label position. (2), this sensitivity is the functional equivalent of label noise, in that it produces seemingly random and unpredictable changes in test set accuracy.

\section{Per-class accuracies on zero-shot datasets}
\label{app:per-class-acc}

\begin{table*}
    \centering{
    \begin{tabular}{@{}lrrrl@{}}
\textbf{Class} & \multicolumn{1}{l}{\textbf{T5}} & \multicolumn{1}{l}{\textbf{UL2}} & \multicolumn{1}{l}{\textbf{GPT}} & \textbf{Conf. Cls.} \\
abstract for patent & 0.99 & 0.97 & 0.95 &  \\
biological formula & 0 & 0 & 0 & chemical \\
book isbn & 1 & 1 & 1 &  \\
book title & 0.06 & 0.52 & 0.28 & journal title \\
cell alternative label & 0.75 & 0.63 & 0.63 & organization \\
chemical & 0.64 & 0.71 & 0.73 & biological formula \\
concept broader term & 1 & 1 & 1 &  \\
disease alternative label & 1 & 1 & 1 &  \\
inchi (international chemical identifier) & 1 & 1 & 1 &  \\
journal issn & 1 & 1 & 1 &  \\
journal title & 1 & 0.97 & 0.98 &  \\
md5 hash & 1 & 1 & 1 &  \\
molecular formula & 0.99 & 1 & 1 &  \\
organization & 0.96 & 0.99 & 0.99 &  \\
patent title & 0.96 & 0.85 & 0.53 & abstract for patent \\
person's first name and middle initials & 0.05 & 0.95 & 0 & names \\
person's full name & 1 & 0.14 & 1 & names \\
person's last name & 0.78 & 0.26 & 0.73 & names \\
smiles (simplified molecular input line entry system) & 0.73 & 0 & 0.21 & chemical \\
taxonomy label & 1 & 1 & 1 & 
\end{tabular}}
    \caption{\sl\textbf{Per class accuracy scores on the Pubchem-20 benchmark. } We observe a high degree of consistency across LLMs in their handling of most classes. LLM classification on PubChem exhibits a high degree of bias in favor of certain semantically distinct classes (person's first name and middle initials, person's last name) over others (person's full name) -- the capacity to distinguish between closely related semantic types zero-shot is limited in current-generation LLMs. Conf. Cls. is a list of classes which were commonly confused with that target class.}
    \label{tab:pubchem-class}
\end{table*}

As a convenient reference, in \Tabref{tab:sotab-class}, \Tabref{tab:d4-class} and \Tabref{tab:pubchem-class} we include per-class accuracies for three of the four zero-shot datasets used in this paper. For details on Amstr, please see our \href{https://github.com/penfever/archetype/}{GitHub repository}.

\section{ArcheType Dataset Examples}

In Fig. \ref{fig:dataset-samples}, we provide examples randomly sampled from each of our new zero-shot benchmarks.

\begin{figure}[th]
    \centering
    \begin{tcolorbox}[colback=green!5!white,colframe=green!75!black]
    \begin{small}
\texttt{\textbf{SOTAB-27}}
 \end{small}
  \begin{small}
 \begin{verbatim}
CREATIVE WORK : "What to Expect When You’re Expecting 
(4th Edition)", "The Better Baby Book: How to Have..."
PERSON: "Otoo, Sharon Dodua", "Danysz, Magda"
PRODUCT: "SKL-200", "1160" \end{verbatim}
 \end{small}
\end{tcolorbox}

\begin{tcolorbox}[colback=red!5!white,colframe=red!75!black]
   \begin{small}
    \texttt{\textbf{D4-20}}
 \end{small}
  \begin{small}
 \begin{verbatim}
EDUCATIONAL ORGANIZATION: "The Global Learning Collab", 
"P.S. 057 Hubert H. Humphrey"
REGION IN THE BRONX: "Bathgate", "Crotona Park East"
CITY AGENCY (FULL): "Mayor's Office of Media 
and Entertainment (MOME)", "Department of Design 
and Construction (DDC)" \end{verbatim}
 \end{small}
\end{tcolorbox}
\begin{tcolorbox}
[colback=blue!5!white,colframe=blue!75!black]
    \begin{small}
    \texttt{\textbf{Amstr-56}}
 \end{small}
  \begin{small}
 \begin{verbatim}
ARTICLE FROM PA: "REDLANDS, Feb. 6.-The Casa Iona hotel 
was last evening the scene..."
NEWSPAPER: "The Nome nugget.", "The Arizona champion." \end{verbatim}
 \end{small}
\end{tcolorbox}
\begin{tcolorbox}
[colback=orange!5!white,colframe=orange!75!black]
    \begin{small}
\textbf{\texttt{Pubchem-20}}
 \end{small}
  \begin{small}
 \begin{verbatim}
MOLECULAR FORMULA: "C10H30Cl4O2Si4", "C43H75NO10S"
DISEASE: "Myofibrillar myopathy, filamin C-related" \end{verbatim}
 \end{small}
\end{tcolorbox}
    \caption{\sl\textbf{ArcheType benchmarks.} \sl  To test the limits of zero-shot CTA, we utilize four zero-shot datasets in our benchmark, three of which are newly introduced with this paper. SOTAB-27 is composed of Schema.org data from different websites.\cite{sotab} D4Tables (D4-20) consists of data from NYC Open Data. AmstrTables (Amstr-56) is composed of newspaper articles, bylines and localization information for content written between 1774 and 1963. PubchemTables (Pubchem-20) is derived from RDF triples consisting of molecular formula, structure, biological activities, safety and toxicity information.}
    \label{fig:dataset-samples}

\end{figure}

\section{Label remapping counts per experiment}

By way of illustrating the high degree of variance in model fidelity to the provided label set, here we report the number of samples remapped per dataset on five randomly selected experiments from our ablation studies.

\begin{small}
\begin{verbatim}
    Amstr-56 (n=2000): 172, 213, 328, 808, 1429
    D4-20 (n=2000): 0, 105, 107, 109, 235
    Pubchem-20 (n=2000): 1, 30, 109, 201, 531
    SOTAB-27 (n=15040): 85, 256, 729, 2133, 3960
\end{verbatim}
\end{small}

Comparing these figures to \Tabref{tab:main-results-zs}, it is apparent there is a positive link between model accuracy and \# of remapped samples.
\end{appendix}

\end{document}